\RequirePackage{fix-cm}

\documentclass[twocolumn]{svjour3}
\smartqed 
\usepackage{graphicx}
\usepackage{times}
\usepackage{epsfig}
\usepackage{color}
\usepackage{pifont}
\usepackage{amsmath}
\usepackage{amssymb}
\usepackage{array}
\usepackage{balance}
\usepackage{multirow}
\usepackage{makecell}
\usepackage{eso-pic}
\usepackage{multirow}
\usepackage{mathrsfs}
\usepackage{hyperref}
\usepackage{tikz}
\usepackage{bbding}
\usepackage{booktabs}
\usepackage{appendix}
\newcommand*{\circled}[1]{\lower.7ex\hbox{\tikz\draw (0pt, 0pt)
    circle (.5em) node {\makebox[1em][c]{\small #1}};}}
\newcommand{\rmnum}[1]{\romannumeral #1}
\newcommand{\Rmnum}[1]{\uppercase\expandafter{\romannumeral #1}}
\newcommand{\tabincell}[2]{\begin{tabular}{@{}#1@{}}#2\end{tabular}}

\newcommand{\xf}[1]{{\color{blue} #1}}

\begin{document}
\begin{sloppypar}

\title{Indoor Obstacle Discovery on Reflective Ground via Monocular Camera}

\author{Feng Xue \and
        Yicong Chang \and 
        Tianxi Wang \and
        Yu Zhou \and 
        Anlong Ming
}

\institute{
Corresponding author : Yu Zhou\\
F. Xue, Y. Chang and A. Ming \at
School of Computer Science, Beijing University of Posts and Telecommunications, Beijing 100876, China \\
\email{\{xuefeng,yicongchang,mal\}@bupt.edu.cn}
\and
Y. Zhou \at
School of Electronic Information and Communications, Huazhong University of Science and Technology, Wuhan 430074, China\\
\email{yuzhou@hust.edu.cn}
\and
Tianxi Wang \at
DiDi Autonomous Driving, Beijing 100094, China\\
\email{wangtianxi@didiglobal.com}
}

\date{Received: 16 March 2022 / Accepted: 30 September 2023}

\maketitle

\begin{abstract}
Visual obstacle discovery is a key step towards autonomous navigation of indoor mobile robots.
Successful solutions have many applications in multiple scenes.
One of the exceptions is the reflective ground.
In this case,
the reflections on the floor resemble the true world,
which confuses the obstacle discovery and leaves navigation unsuccessful.
We argue that the key to this problem lies in obtaining discriminative features for reflections and obstacles.
Note that obstacle and reflection can be separated by the ground plane in 3D space.
With this observation,
we firstly introduce a pre-calibration based ground detection scheme that uses robot motion to predict the ground plane.
Due to the immunity of robot motion to reflection,
this scheme avoids failed ground detection caused by reflection.
Given the detected ground,
we design a ground-pixel parallax to describe the location of a pixel relative to the ground.
Based on this,
a unified appearance-geometry feature representation is proposed to describe objects inside rectangular boxes.
Eventually,
based on segmenting by detection framework,
an appearance-geometry fusion regressor is designed to utilize the proposed feature to discover the obstacles.
It also prevents our model from concentrating too much on parts of obstacles instead of whole obstacles.
For evaluation,
we introduce a new dataset for Obstacle on Reflective Ground (ORG),
which comprises 15 scenes with various ground reflections,
a total of more than 200 image sequences and 3400 RGB images.
The pixel-wise annotations of ground and obstacle provide a comparison to our method and other methods.
By reducing the misdetection of the reflection,
the proposed approach outperforms others.
The source code and the dataset will be available at \textit{\url{https://github.com/XuefengBUPT/IndoorObstacleDiscovery-RG}}.

\end{abstract}

\section{Introduction}
\label{sec:introduction}
\begin{figure}
	\centering
	\includegraphics[width=1\linewidth]{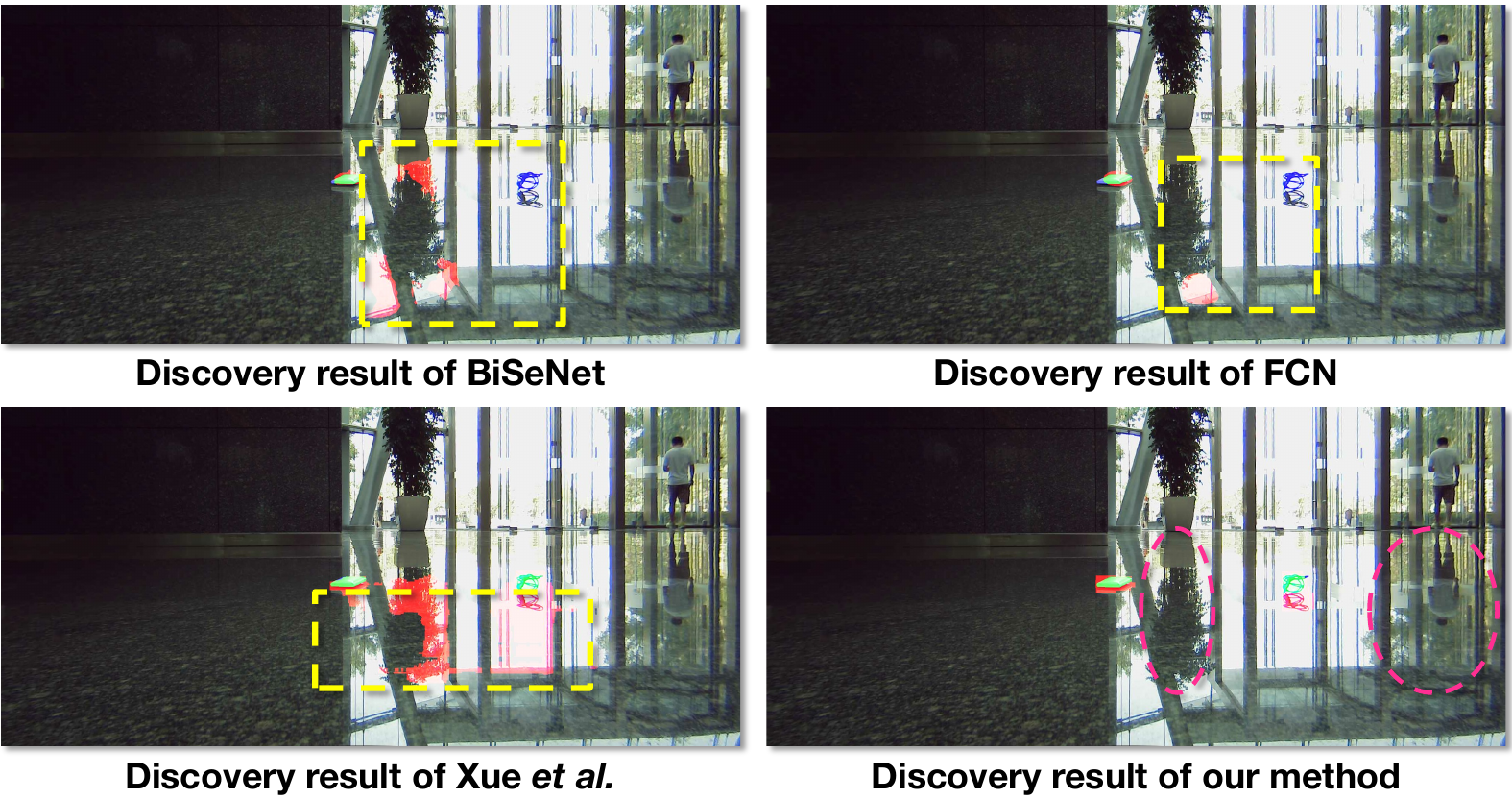}
		\caption{
		Results of BiSeNet \cite{BiSeNet}, FCN \cite{FCN}, Xue \emph{et al}. \cite{ICRA} and our method on an exemplary scene of the proposed dataset.
		True positives are marked in green,
		red for false positives,
		blue for false negatives.
		Yellow boxes mark the mis-classified pixels.
		Magenta circles indicate the reflection.}
	\label{fig:intro}
\end{figure}

In indoor environments,
obstacles, e.g., charging cable, key chain, and wallet, etc.,
endanger mobile robots by tangling wheels or causing the robot to overturn.
However,
the 2D LiDAR commonly used for navigation only perceives obstacles with large heights,
while obstacles lower than LiDAR are hard to be perceived.
Hence,
as a potentially effective way,
cost-effective cameras are usually used in many researches \cite{JinICRA,Markov,EM} for discovering these hazards.
While most existing approaches are mainly applied in environments with texture-less and non-reflective floor,
and lack discussion on the indoor scene laying reflective floor.
In practical scenarios,
the floor with a mirror surface is prevalent and brings a challenge to obstacle discovery.
In the Field of View (FOV) of a robot,
the floor reflects real-world objects that we call unreal objects (UOs),
% the floor presents the reflections of real-world objects that we call unreal objects (UOs),
e.g. the reflections of plant, gate, and person in Fig. \ref{fig:intro}.
UOs generally have complex textures and surround obstacles,
making them even harder to detect using classic depth sensors (see Appendix \ref{secC}) or eliminate with reflection removal algorithm (see Appendix \ref{secB}).
This makes it challenging for discovery models to distinguish the obstacles from the floor,
ultimately leads to confusion in robot navigation.

As we know,
there are currently no obstacle discovery methods investigating this task.
Thus,
in the following paragraphs,
we discuss the feasibility of using the existing obstacle discovery methods on reflective ground.
Overall,
they can be mainly assigned to three groups.

\begin{figure*}
	\centering
	\includegraphics[width=0.96\linewidth]{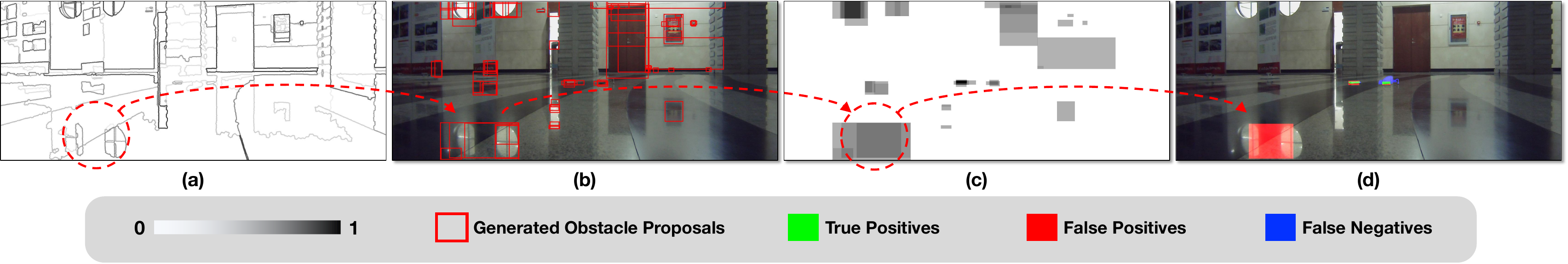}
		\caption{
		The result obtained by previous method \cite{ICRA}.
		(a) the occlusion edge map.
		(b) the region proposals with high score (marked in red boxes).
		(c) the obstacle-occupied probability map constructed by these high-score boxes.
		(d) the final obstacle masks (using threshold 0.49 to segment the obstacles).
		In the grayscale images (a)(c),
		the value of each pixel ranges from 0 to 1.
		The darker it is,
		the closer it is to 1.}
	\label{fig:rw}
\end{figure*}

\noindent
\textbf{Stereo-based conventional methods} \cite{LAF,2011Stereo} reconstruct 3D scenes by using a stereo camera,
and classify the 3D points higher than the ground plane as obstacle.
However,
the accuracy of obstacle discovery highly depends on the quality of 3D reconstruction that is easily affected by the mirror reflection.
Thus, obstacles cannot be distinguished from UOs by using the unreliable 3D information.

\noindent
\textbf{Monocular-based conventional methods} \cite{JinICRA,EM,JinICIP,Lin} first estimate a ground plane by registered feature point pairs \cite{ICP,Chen2009vc} between consecutive monocular images.
Then, they judge whether a pixel is coplanar with the estimated ground.
The pixels that are not coplanar with the ground indicate obstacles.
However,
the feature points of reflection dominate the ground area,
and their 3D information is totally different from the ground plane,
leading to the mis-detection of ground.
For this reason,
these solutions cannot be directly used to distinguish obstacles from UOs.

\noindent
\textbf{Learning-based methods} \cite{Markov,ICRA,UON,MergeNet,tip,9134735,Lis} train models to classify pixels or region proposals \cite{Ghodrati2017ve,Li2019wt} in a single image into obstacle and non-obstacle.
However,
these methods do not capture the difference between reflections and the real world.
Thus, in the face of complex, diverse and unseen ground textures,
it is difficult for them to avoid mis-detecting the ground as obstacles.
Fig.\ref{fig:intro} depicts the prediction of two segmentation models and the baseline.
These methods mistake the reflection of plant and gate for obstacles,
and miss the real obstacles.

Although the use of homography for obstacle detection is not really new \cite{JinICIP,JinICRA,EM},
we found that homography is good for expressing the difference between UOs and obstacles because different planes,
which is rarely discussed in previous works.
To fully leverage such a key characteristic,
we first follow \cite{ICRA} to gain the candidate proposals enclosing objects,
and then construct a unified appearance-geometry feature representation to express and re-score the candidate proposals.
In detail,
the homography of ground is firstly estimated for feature construction.
To avoid the failed ground detection \cite{JinICRA,JinICIP} on the reflective floor,
a pre-calibration based ground detection scheme is introduced.
This scheme uses robot motion,
instead of registered feature point pairs \cite{JinICRA},
to figure the ground homography.
Thus,
it avoids failed detection caused by reflection.
Secondly,
we take the occlusion edge point \cite{IS} of the scene as the key point for feature extraction,
and propose a ground-pixel parallax.
It measures the homography difference between an occlusion edge point and the detected ground,
thus is able to express whether this point is above or below the ground.
Then,
as a key discriminative feature,
the parallax incorporates with appearance features \cite{ICRA} to form a unified appearance-geometry feature representation for region proposals.
Finally,
an appearance-geometry fusion model (AGFM) is designed to re-score all proposals.
In AGFM,
we carefully study the effects of fast-moving objects on features,
and adaptively use the geometric and appearance features to eliminate the effects of fast-moving.
Besides,
to segment obstacles more accurately,
AGFM utilizes a well-designed weight-decayed probability generation scheme to gain an obstacle-occupied probability map.
It reduces weights of low-rank obstacle proposals to avoid concentrating too much on parts of obstacles.

To evaluate the effectiveness of our method,
a dataset named Obstacle on Reflective Ground (ORG) is introduced,
which is the first dataset focusing on discovering obstacles on reflective floor,
as far as we know.
Inspired by the obstacle discovery task in other scenarios \cite{ICRA,MergeNet,Hua_ICCVW},
several segmentation methods \cite{FCN,DeepLab,BiSeNet,ENet,ICRA} are employed to compare with our method.
The experimental results prove that the presented approach significantly alleviates false positives and false negatives compared with other methods,
and is robust against the noise of robot motion and motion blur.

To our knowledge,
our method is the first stab to discover obstacle on reflective ground.
The key insights lie in:
\begin{itemize}
\item
A pre-calibration based ground detection scheme is introduced,
which uses robot motion that is immune to reflection,
thus avoiding detection failure.
\item
A ground-pixel parallax is introduced to form a unified appearance-geometry feature representation for region proposals together with appearance feature.
\item
An appearance-geometry fusion model (AGFM) is proposed to locate obstacle.
It also avoids the performance degradation caused by fast-moving objects and proposals too concentrated on parts of obstacles.
\item An Obstacle on Reflective Ground (ORG) dataset is proposed,
which contains 15 challenging scenarios, and 200 sequences.
Compared to other segmentation models,
our approach achieves a better performance on this dataset.
\end{itemize}

\section{Related Work}
Since our method is related to the general obstacle discovery \cite{ICRA,tip},
occlusion edge,
proposal extraction,
and the ground homography.
Brief introductions for them are given.

\begin{figure*}
	\centering
	\includegraphics[width=1\linewidth]{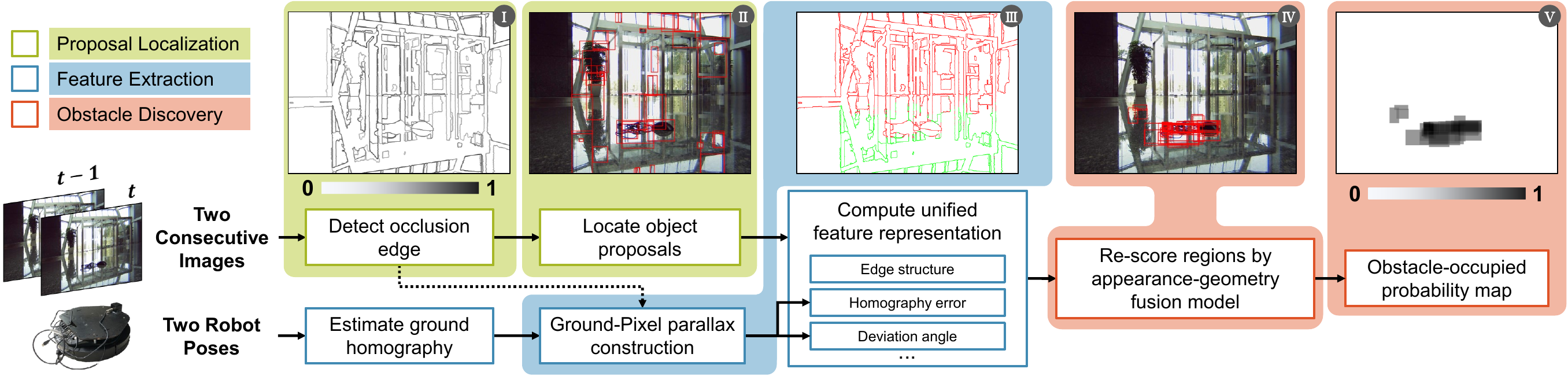}
		\caption{The pipeline of our method.
		The inputs are consecutive RGB images and the robot pose corresponding to the two images.
		\Rmnum{1}-\Rmnum{4} are the byproducts.
		\Rmnum{5} is the output,
		an obstacle-occupied probability map.
		In \Rmnum{1} and \Rmnum{5},
		the value of each pixel ranges from 0 to 1.
		The darker it is,
		the closer it is to 1.
		In \Rmnum{3},
		the points are divided into two types by setting threshold:
		the red points are above the ground,
		the green points are below the ground.
		}
	\label{fig:pipeline}
\end{figure*}

\subsection{The Method Segmenting Obstacle by Detection}
Xue et al. \cite{ICRA} propose to segment the obstacle by detection.
In this pipeline,
each region proposal obtained by \cite{OLP} is expressed by a feature vector,
which is used to train the random forest to detect the obstacle from proposals.
Finally, all obstacle proposals are used to construct an obstacle-occupied probability map which is used to determine the obstacle pixels.
Due to locating obstacles by the confidence of numerous proposals, instead of one proposal,
this method achieves an approximate performance of the Convolutional Neural Networks (CNN) based methods \cite{MergeNet,UON}.

Unfortunately,
the UOs are usually so complicated that the existing features cannot express the difference between them and obstacles,
e.g. the occlusion edge in Fig.\ref{fig:rw}(a).
Hence,
the UOs and obstacles are captured indiscriminately (see Fig.\ref{fig:rw}(b)).
Eventually,
the UOs obtain high confidence and are mis-detected as obstacles,
as shown in Fig.\ref{fig:rw}(c)(d).
This phenomenon accounts for a large proportion of indoor scenarios.
Thus,
our method aims to distinguish the proposals of UOs and obstacles by appearance-geometry features,
boosting the performance of obstacle discovery.

\subsection{Occlusion Edge and Region Proposal}
\label{sec:OERP}
Occlusion edge \cite{Hoiem2011te,OFNet,ICIP15} locates the pixels indicating depth discontinuity between objects and background.
Due to exploiting surface cue,
it generally has a more robust confidence in the object contour than the typical edge cues \cite{Xie2017tc,gPb,Lindeberg1998tl}.
By considering the occlusion edge points inside a bounding box,
object-level proposal (OLP) \cite{OLP} models an occlusion-based objectness score to measure the probability that the bounding box contains object.
The high-score bounding boxes are retained from densely sampled sliding windows,
and are considered as the candidate proposals.

In this paper,
we firstly employ \cite{IS} to detect the occlusion edge from the scene (see Fig.\ref{fig:pipeline} \Rmnum{1}),
and then utilize OLP \cite{OLP} to extract a set of candidate obstacle proposals (see Fig.\ref{fig:pipeline} \Rmnum{2}).
The resulting occlusion edge and proposals are taken as the inputs of our method.

\subsection{Homography of Planar Surface}
With the representation in \cite{JinICRA,Markov,JinICIP},
for a set of the point pairs $\{\mathrm{x}_i\leftrightarrow \mathrm{x}'_i\}$ from two images,
if the points $\{\mathrm{x}_i\}$ are coplanar,
there is a homography matrix $\mathrm{H}\in\mathbb{R}^{3\times3}$:
\begin{equation}
    \mathrm{x}_i = \mathrm{H} \mathrm{x}'_i
\end{equation}
where $\mathrm{x}_i,\mathrm{x}'_i\in\mathbb{R}^{3\times1}$ are the homogeneous image coordinates.
Since the matrix $\mathrm{H}$ has eight degrees of freedom \cite{JinICIP,MV},
a minimum of four non-degenerate point correspondences are required to determine $\mathrm{H}$.

By searching for the coplanar points on the ground and registering them between two frames,
several methods \cite{EM,Panahandeh} estimate the homography matrix to represent the ground plane.
However,
in the case of reflective ground,
the complex UOs produce a lot of feature point pairs that are the outliers to the ground homography.
Evidently,
detecting the ground plane from point pairs with numerous outliers is difficult.
Even,
this phenomenon easily results in the virtual plane problem \cite{JinICRA},
namely, detecting a plane that does not correspond to the physical ground plane.
In this paper,
we utilize the robot motion,
instead of the point pairs,
to calculate the ground homography.
Since the robot motion is independent of ground appearance,
the proposed ground detector is immune to reflection.

\section{Method}
\subsection{Overview}
\label{sec:overview}
Given an image sequence $\{I^1, I^2,..., I^t\}$ acquired by the robot,
where $t\in\mathbb{N+}$ is the current time,
our approach takes two frames $I^t$ and $I^{t-q}$ as inputs and generates a probability map $P$.
Each element in the map $P$ represents the presence of obstacles in image $I^t$ and $q\in\mathbb{N+}$ is the time interval.
% in which each element indicates the obstacles in image $I^t$ and $q\in\mathbb{N+}$ is the time interval.
% Specifically,
% we consider a single previous frame $I^{t-q}$ in this paper,
% where $q\in\mathbb{N+}$ is the time interval. 
To simplify the representation,
we default $q$ to 1.

% Given an image sequence $\{I^1, I^2,..., I^t\}$ acquired by the robot,
% where $t\in\mathbb{N+}$ is the time,
% the goal of our approach is to predict a probability map $P$, 
% in which each element indicates the obstacles in image $I^t$.
% Specifically,
% we consider a single previous frame $I^{t-q}$ in this paper,
% where $q\in\mathbb{N+}$ is the time interval. 
% To simplify the representation,
% we default $q$ to 1.

The pipeline is shown in Fig.\ref{fig:pipeline}.
We extract the occlusion edge map and the candidate proposals from image $I^t$,
as mentioned in Sec. \ref{sec:OERP}.
To reduce the computation,
we only preserve $\tau_e$ percent occlusion edge points with the top response.
The candidates with the highest objectness score are denoted as $\mathcal{B} = \{b^t_j|j\in\{1,2,...,J\}\}$,
as the red rectangles shown in Fig. \ref{fig:pipeline} \Rmnum{2}.
However, through the projection from the 3D space to the 2D image plane,
the detailed 3D information is lost in the 2D image.
Thus, $\mathcal{B}$ contains many false-positive obstacle proposals caused by the reflective ground.
A key observation of our approach is that different planes in the 3D space satisfy different homography, 
thus, aiming to distinguish the real obstacles and the UOs, 
we leverage the discriminative characteristic of the homography,   
construct a unified appearance-geometry feature representation to express and re-score the obstacle proposals in $\mathcal{B}$.

\begin{figure*}
	\centering
	\includegraphics[width=1\linewidth]{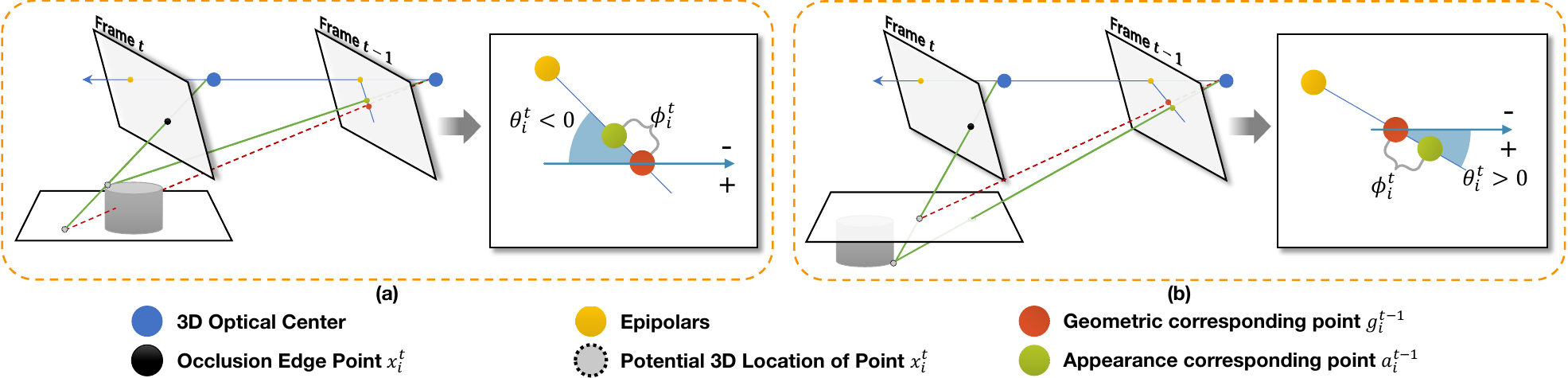}
		\caption{
		(a) the case that the observed point is above the ground.
		(b) the case that the observed point is below the ground.
		For each case,
		the points are zoomed in to the right-side images to obtain a clear view.
		}
	\label{fig:rw2}
\end{figure*}

More specifically, 
taking the occlusion edge points $\mathcal{P}^{t} = \{x^{t}_{1}, x^{t}_{2},..., x^{t}_{n}\}$
as the key point for feature extraction,
the Lucas-Kanade(LK) optical flow approach \cite{LK} is employed to track the occlusion edge point $x^t_i \in \mathcal{P}^t$ from image $I^t$ to $I^{t-1}$,
capturing the appearance corresponding point $a^{t-1}_i$ on image $I^{t-1}$.
In addition, we leverage the ground plane as the base reference plane, 
and compute the homography $\mathrm{H}$ of the ground plane (see Sec 3.2).
By using $H$, we find the geometric corresponding point $g_{i}^{t-1}=\mathrm{H}x^t_i$ of $x_{i}^{t} \in \mathcal{P}^{t}$ in $I^{t-1}$. 
Furthermore, we define the parallax of the points by considering the 
relationship of $a^{t-1}$ and $g_{i}^{t-1}$ (see Sec \ref{sec:psd}). 
As shown in Fig. 4, 
such a parallax reflects a significant difference when the obstacle exists above or under the ground,
and hence we leverage such a difference as a key discriminative feature to distinguish the UOs. 
Therefore, we gather the parallax values of the occlusion edge points inside a region proposal,
and then express a proposal by jointly considering both the appearance cues and the geometric cues (see Sec \ref{sec:feat}).
An appearance-geometry fusion model (AGFM) is trained to re-score all region proposals (see Sec 3.5).
Multiple proposals with the highest confidence are employed to form an obstacle-occupied probability map. 
Compared with the existing approaches, 
our approach takes the intrinsic geometric cue of the proposals into consideration and hence alleviates the false-positive obstacle proposals caused by the reflection of the ground efficiently.

\subsection{Ground Plane Detection via Pre-calibration}
\label{sec:homo}

The homography of the ground plane between both consecutive frames is a basic component for the geometric representation of proposal.
However,
the reflection obstructs the ground detection.
To detect the ground without the effect from reflection,
we first calibrate the ground parameter offline.
In detail,
since the camera is mounted on the robot platform that moves on a planar floor,
the ground is indicated as a constant vector $\pi=\{\mathbf{n}^\mathsf{T},\mathbf{d}\}$,
and the camera moves on a plane that satisfies $\mathbf{n}^\mathsf{T}X+\mathbf{d}=0$,
where $\mathbf{n}^\mathsf{T}\in \mathbb{R}^{3\times1}$ denotes the ground normal.
The camera height $\mathbf{d}$ can be measured directly,
and the normal $\mathbf{n}^\mathsf{T}$ is calibrated as follows:
\begin{enumerate}
    \item Take two images by the camera in two different poses of robot.
    \item Manually mark a few ($\ge 4$) ground points pairs.
    $\{x_i\}$ and $\{x'_i\}$ denote the ground points of the two images,
    which is one-to-one correspondences: $\{x_i\leftrightarrow x'_i \}$.
    \item Compute a homography $H$ by these correspondences: $x'_i = Hx_i$.
    \item Decompose the homography $H$ by the decomposing method \cite{DecompH} to obtain a proper ground normal $\mathbf{n}^\mathsf{T}$.
\end{enumerate}

Based on the pre-calibrated ground $\pi$ and the robot pose,
the ground homography can be determined online.
Specifically,
we firstly utilize the odometer-camera calibration approach \cite{CalibUrl,Heng2013CamOdoCal} to obtain the transformation matrix from the robot's wheel odometer to the camera.
Then,
given the robot poses from the odometer at time $t$ and $t-1$,
the poses of the camera at time $t$ and $t-1$ are figured by transformation,
which are denoted as $T^t$ and $T^{t-1}$,
where $T^t = [R^t|C^t](R^t\in \mathbb{R}^{3\times3},C^t\in \mathbb{R}^{3\times1})$.
Referring to \cite{JinICIP,MV},
the homography of ground $\pi$ can be directly calculated by the formulation in \cite{MV}:
\begin{equation}
    \mathrm{H}=K(\Delta R-\Delta C\mathbf{n}^\mathsf{T}/\mathbf{d})K^{-1}
    \label{equ:homography}
\end{equation}
where $K\in \mathbb{R}^{3\times3}$ is the intrinsic matrix of the camera,
and $\Delta R=R^{t-1}(R^t)^{-1}$ is the relative rotation of the camera from $T^t$ to $T^{t-1}$,
$\Delta C=C^{t-1}-C^t$ corresponds to the relative translation.
Since the ground vector $\mathbf{n}^\mathsf{T}$ is pre-calibrated,
and the camera motion $[\Delta R|\Delta C]$ is independent of the pattern inside the ground area,
Equation 2 avoids false ground detection when working on reflection floor.
Thus, the proposed scheme is suitable for reflective ground.

\subsection{Ground-Pixel Parallax of Occlusion Edge Point}
\label{sec:psd}
By modeling the contrastive property between point and ground plane,
we propose ground-pixel parallax to judge whether an occlusion edge point is real or reflection.
With the representation in Sec. \ref{sec:overview},
the observed occlusion edge points are denoted as $\mathcal{P}^{t} = \{x^{t}_{1}, x^{t}_{2},..., x^{t}_{n}\}$.
For each point inside set $\mathcal{P}^{t}$,
the point's appearance and geometry corresponding points in frame $t-1$ are denoted as $a^{t-1}_i$ and $g^{t-1}_i$,
as shown by the green and red circles in Fig.\ref{fig:rw2}.
In addition,
we denote the epipoles of frames $I^t, I^{t-1}$ as $e^t$ and $e^{t-1}$,
namely, the yellow circles in Fig.\ref{fig:rw2}.
Since $g^{t-1}_i$ is determined by the ground homography $\mathrm{H}$,
it can be considered as the projection of a 3D ground point.
Hence,
the parallax of point $x^t_i$ to the ground is defined as the difference between the appearance corresponding point $a^{t-1}_i$ and the geometry corresponding point $g^{t-1}_i$:
\begin{equation}
\mathbf{p}^t_i = a^{t-1}_i - g^{t-1}_i
\label{eq:pp}
\end{equation}
According to \cite{MV},
since the three points $g^{t-1}_i$, $e^{t-1}$, and $a^{t-1}_i$ are collinear,
the vector $\mathbf{p}^t_i$ can also be extended as:
\begin{equation}
\mathbf{p}^t_i = a^{t-1}_i - g^{t-1}_i = \rho(g^{t-1}_i-e^{t-1})
\label{eq:pp2}
\end{equation}
where $\rho$ is a scalar representing the deviation relative to the ground.
It indicates which side of ground the real 3D location of point $x^t_i$ is in the following way:
\begin{itemize}
\item If $\rho=0$,
the real 3D location is coplanar with the ground.
\item If $\rho<0$,
the real 3D location is above the ground,
as depicted in Fig.\ref{fig:rw2} (a).
\item If $\rho>0$,
the real 3D location is under the ground,
as depicted in Fig.\ref{fig:rw2} (b).
\end{itemize}
Observably,
vector $\mathbf{p}^t_i$ reveals that point $x^t_i$ is above or below the ground.
The proofs of Equation \ref{eq:pp2} and detailed technical discussion are included in the supplementary materials.

\subsection{Appearance-Geometry Feature Representation}
\label{sec:feat}
For each proposal $b^t_j$ in $\mathcal{B}$,
we propose a unified appearance-geometry feature representation,
which has two parts,
as shown in Table \ref{table:features}.
Inspired by another obstacle detector \cite{ICRA},
the first part contains edge cue, pseudodistance, objectness score, and color cue.
For the second part,
we gather the parallaxes of occlusion edge points inside the proposal as feature.
As the ground-pixel parallax $\mathbf{p}^t_i$ cannot be normalized and is easily affected by failed optical flow tracking,
we decompose it into two features: homography error and deviation angle.

\noindent
\textbf{Region-level Homography Error:}
To express the relative distance of a proposal $b^t_j$ to the ground,
we define the homography error of occlusion edge points $x^{t}_i$ enclosed by $b^t_j$:
$\phi_i^t = \|a^{t-1}_i-g^{t-1}_i\|$,
as shown in Fig.\ref{fig:rw2}.
According to \cite{Markov,JinICRA,JinICIP},
it depicts the 2D deviation from occlusion edge point $x^{t}_i$ to the ground.
For proposal $b^t_j$,
the region-level homography error is defined as the mean error of occlusion edge points $x^{t}_i$ enclosed by proposal $b^t_j$:
\begin{equation}
    \Phi_j^t = 1/N^t_j \sum\nolimits_{x^t_i\in b^t_j}\phi_i^t
\end{equation}
where $N^t_j$ is the number of occlusion edge points in the proposal $b^t_j$.
The region-level homography error indicates how far the proposal $b^t_j$ is from the ground.

\begin{table}[!tp]
	\begin{center}
		\caption{
			The appearance-geometry feature representation of region proposal.
		}
		\begin{tabular}{|c |c | c|}
\hline
			{\bf Part} &{\bf Category} & {\bf Feature name}\\
			\hline
			\multirow{13}{*}{{\it Appearance}} & \multirow{4}{*}{{\it Edge cue}} & max edge response \\
			& & proportion of most response\\
			& & average edge response \\
			& & average edge response in inner ring \\
			\cline{2-3}
			&\multirow{4}{*}{{\it \tabincell{c}{Pseudo\\distance}}} & normalized area \\
			& & aspect ratio \\
			& & X,Y coordinate of the region center \\
			& & width, height \\
			\cline{2-3}
			& {\it Objectness} & occlusion-based objectness \\
			\cline{2-3}
			& \multirow{2}{*}{{\it Color}} &  color deviations in the H,S,V channel \\
			& & color contrasts of the H,S,V channel \\
			\hline
			\multirow{2}{*}{{\it  Geometry}} &\multirow{2}{*}{{\it Parallax}} &  homography error \\
			& & deviation angle \\
\hline
		\end{tabular}
		\label{table:features}
	\end{center}
\end{table}

\noindent
\textbf{Region-level Deviation Angle:}
To express which side of the plane $\pi$ the observed 3D point is,
we define the deviation angle as the angle of point $a^{t-1}_i$ in the polar coordinate system centered on point $g^{t-1}_i$,
which is formulated as:
\begin{equation}
    \theta^t_i = \gamma arcsin(\frac{[a_i^{t-1}]_2 - [g^{t-1}_i]_2}{\phi_i^t})
\end{equation}
where $[ . ]_2$ denotes taking y-value of a pixel.
$\gamma$ is a parameter,
which is $+1$ when the robot moves forward,
conversely $-1$.
The region-level deviation angle of proposal $b_j^t$ is stated as:
\begin{equation}
    \Theta_j^t = 1/N^t_j \sum\nolimits_{x^t_i\in b^t_j}\theta_i^t
\end{equation}
Although it seems feasible to use $[a_i^{t-1}]_2 - [g^{t-1}_i]_2$ directly to distinguish the proposals of obstacles and UOs,
the deviation angle $\theta_i^t$ is more suitable.
The reason is that $\theta_i^t$ has a fixed value range and is not easily affected by sharply changing noise points,
which is further proved by the experiments in Sec \ref{sec:ablafeat}.

\noindent
\textbf{Region-level Feature Confidence:}
However,
in a hands-on environment,
robots might move a lot suddenly,
causing the objects move fast between frames.
Due to the high sensitiveness of the optical flow tracking to fast moving,
the point $x_i^{t-1}$ is located inaccurately at times.
Thus, feature confidence is necessary.
Specifically,
the forward-backward error \cite{Kalal2010Forward},
indicating the distance between the original point and its position after the forward and backward optical flow tracking,
is employed to calculate feature confidence.
Assuming that edge point $\mathbf{x}^{t}_i$ is the backward optical flow point of $a^{t-1}_i$ from images $I^{t-1}$ to $I^t$,
the forward-backward error of point $x^{t}_i$ is formulated as the Euclidean distance from $\mathbf{x}^{t}_i$ to $x^{t}_i$: $\lambda_i^t = \|x^{t}_i-\mathbf{x}^{t}_i\|$.
Thus, the feature confidence of proposal $b^t_j$ is stated as:
\begin{equation}
    \Lambda_j^t = 1/\sqrt{(1/N^t_j\sum\nolimits_{x^t_i\in b^t_j}\lambda_i^t)}
\end{equation}
If the point $a^{t-1}_i$ calculated by tracking $x^{t}_i$ is inaccurate,
the backward-tracking point $\mathbf{x}^{t}_i$ would be far away from the original point $x^{t}_i$,
which leads to a large distance $\lambda_i^t$.
In this case,
the confidences of the proposals containing point $x^{t}_i$ are decreased.

\begin{figure*}
	\centering
	\includegraphics[width=1\linewidth]{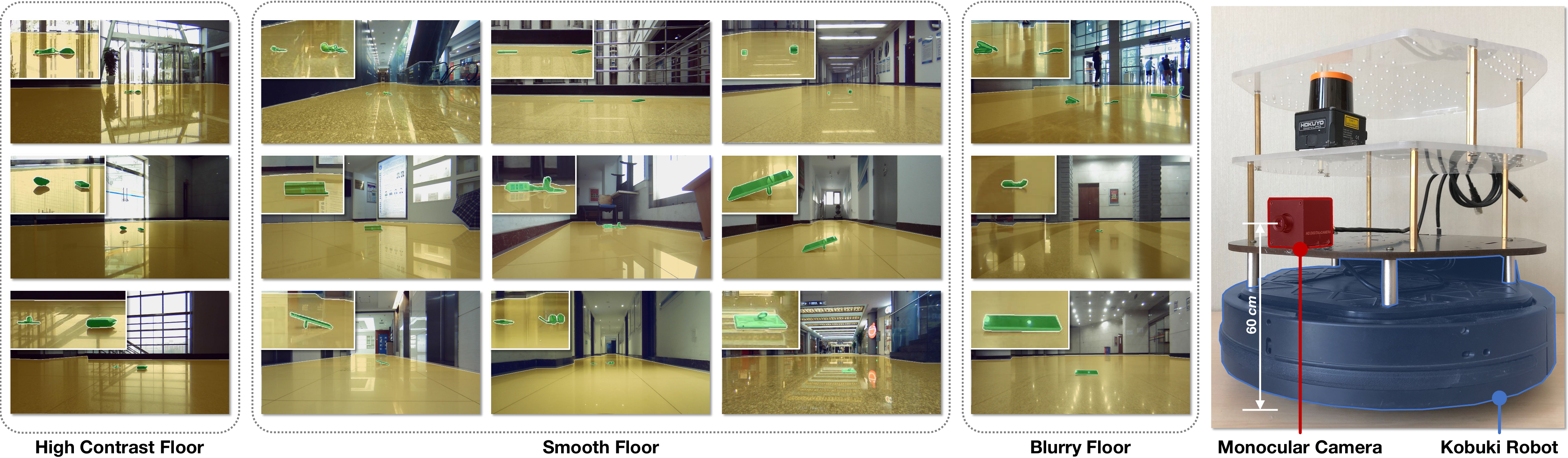}
		\caption{
		Exampler images and pixel-level annotations taken from the proposed dataset.
		The floor is marked in yellow,
		and green for obstacle.
		The images are zoomed in to clearly show these obstacles.
		The right-side image exhibits our platform,
		a Kobuki robot that provides the odometer data.}
	\label{fig:exam}
\vspace{-10pt}
\end{figure*}

\begin{figure}
	\centering
	\includegraphics[width=1\linewidth]{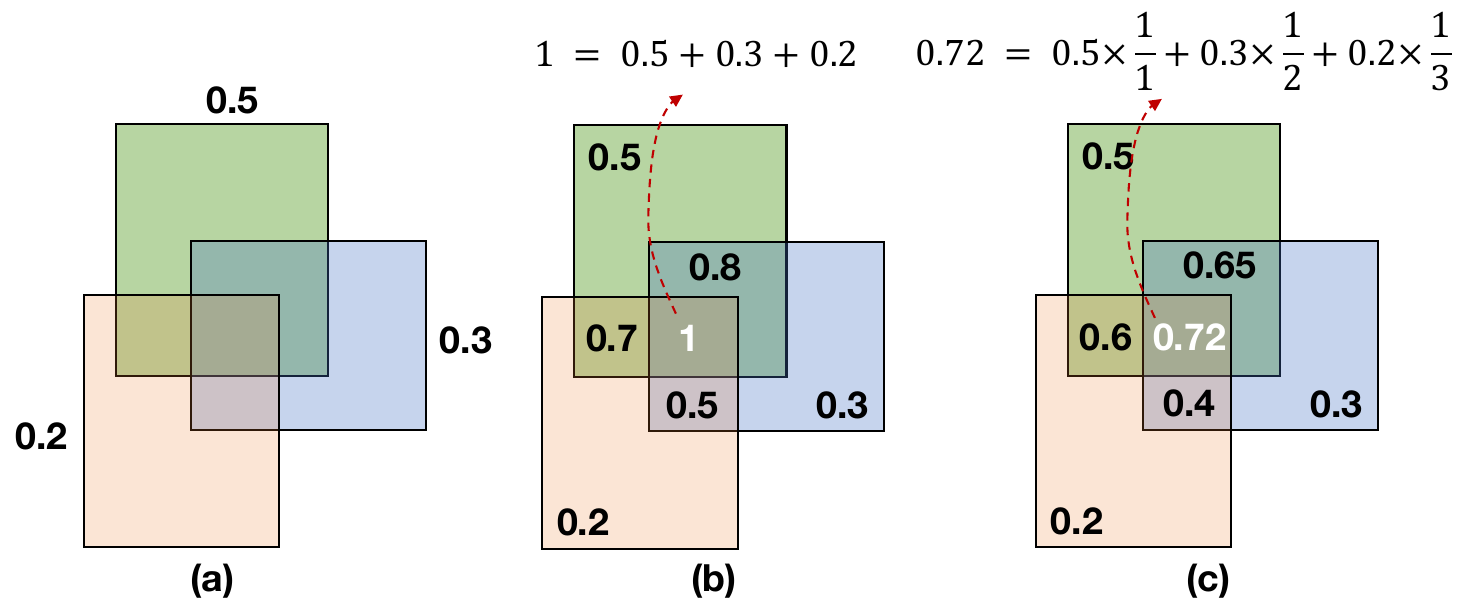}
		\caption{The illustration of weight-decayed probability generation.
		(a) three bounding boxes and their confidences.
		(b) generated probabilities without decayed weights.
		(c) proposed generation scheme.}
	\label{fig:pm}
\end{figure}

\subsection{Appearance-Geometry Fusion Model}
\label{sec:Model}
To locate the obstacles,
we re-score the proposals by a joint model, i.e., appearance-geometry fusion model (AGFM),
that consists of two parts:
The first appearance-geometry regressor (AGR) distinguishes UO and obstacle which fully uses the proposed features.
The second appearance regressor (AR) handles proposals affected by fast motion.

\subsubsection{Model Structure}
Our AGFM employs a random forest structure \cite{DF} to achieve an accuracy-efficiency trade-off.
In detail,
the regressor $\text{AGR}=\{f^{ag}_k|k=1,...,K^{ag}\}$ consists of $K^{ag}$ decision trees,
and each one contains several internal nodes and leaf nodes.
With the same structure,
the regressor $\text{AR}$ contains $K^{a}$ trees,
and $f^{a}_k$ for each tree in $\text{AR}$.
In the training phase,
the extracted proposals are taken as training samples.
Each internal node selects a feature and splits the samples into two parts,
each leaf node stores the mean label of proposals reaching on it during training.
In the inference phase,
by inputting the proposals to root nodes of these trees,
each proposal is separated by internal nodes,
and passed to the left or right pathway until a leaf node is reached.

\subsubsection{Training Data}
For our AGFM,
all the samples, i.e., the proposals $\{b_j^t\}$, are predefined as (\rmnum{1}) floor or (\rmnum{2}) background.
The proposal whose more than 40\% pixels are occupied by floor or obstacle is marked as floor, otherwise background. 
Similar to \cite{ICRA},
only the proposals of the floor are selected for training.
In terms of the features,
as depicted in the last section,
each training sample of AGR is represented by the unified features,
that is,
a sample $b^t_j$ corresponds to 19-dimensional feature vector $v^t_j \in \mathbb{R}^{19}$,
as shown in Table \ref{table:features}.
The detailed formulation of each feature channel can be found in Appendix \ref{secD}.
In contrast to AGR,
the sample of AR is represented by the first 17 dimensions of $v^t_j$,
i.e., denoted as $\bar{v}^t_j \in \mathbb{R}^{17}$.
With the training samples mentioned above,
AGR and AR are trained to regress the Intersection over Union (IoU) between a proposal and obstacle segmentation.
% With the training samples mentioned above,
% AGR and AR are trained to regress the overlap between a proposal and obstacle.

\subsubsection{Prediction}
Following the inference process of decision tree \cite{DF},
each decision tree inputs the 19-dimensional feature vector $v^t_j$,
and outputs the predicted IoU between proposal $b^t_j$ and obstacle.
The outputs of the two regressors are stated as:
\begin{equation}
\begin{split}
\text{AGR}(b_j^t) = 1/K^{ag}\sum\nolimits_{i=1}^{K^{ag}}f^{ag}_i(v_j^t) \\
\text{AR}(b_j^t) = 1/K^{a}\sum\nolimits_{i=1}^{K^{a}}f^{a}_i(\bar{v}_j^t)
\end{split}
\end{equation}
Since some objects move fast in the view of robot,
the failed optical flow tracking leads to a low feature confidence $\Lambda_j^t$.
Thus, confidence $\Lambda^t_j$ indicates if an object moves fast in the field view of robot.
Based on this, the entire prediction is formulated as follows:
\begin{equation}
    F(b_j^t) = 
    \begin{cases}
	\text{AGR}(b_j^t), &\text{if } \Lambda_j^t < \tau_{gc}\\
	\text{AR}(b_j^t), &\text{otherwise}
	\end{cases}
\end{equation}
where $\tau_{gc}$ represents the confidence threshold used to select between AR or AGR for scoring bounding boxes.
The geometry features express the difference between obstacle and UO,
which accordingly helps AGFM to limit the score of the UO.
Furthermore, since the appearance features are independent of the fast motion,
AR makes up for the inaccuracy of AGR in fast motion.

\subsubsection{Weight-decayed Scheme for Obstacle-occupied Map}
After scoring all proposals by AGFM,
the scores of top $\tau_b$ proposals,
i.e., $\{F(b_j^t)|b_j^t\in \hat{\mathcal{B}}\}$,
are accumulated to construct an obstacle-occupied probability map,
where $\hat{\mathcal{B}}$ denotes the set of top $\tau_b$ proposals.
However,
the previous methods \cite{ICRA,tip} often obtain a large number of low-score proposals in a similar location,
even if using Non-Maximum Suppression (NMS) to remove many low-score proposals.
This issue makes the probability map concentrates too much on parts of obstacles,
instead of whole obstacles.

To segment obstacles as completely as possible,
we propose a weight-decayed generation scheme for obstacle-occupied probability map.
In more detail,
assuming that the proposals covering a certain pixel $p$ are sorted in descending order of score,
and the numerical order of the $j$-th proposal is represented as $r^p_j$,
each proposal has a weight that is an reciprocal of order $r^p_j$.
The obstacle-occupied probability of pixel $p$ is formulated as:
\begin{equation}
P(pixel(p)) =
\frac{1}{\mathcal{N}^P} \sum\nolimits_{b_j^t\in \hat{\mathcal{B}}} \mathbf{1}(p, b_j^t) \times F(b_j^t) \times \frac{1}{r^p_j}
\end{equation}
where $pixel(p)$ denotes the coordinate of pixel $p$.
$\frac{1}{\mathcal{N}^P}$ denotes the normalization term.
$\mathbf{1}(p, b_j^t)$ is an indicator. If $p \in b_j^t$, then $\mathbf{1}(p, b_j^t)=1$. Otherwise, $\mathbf{1}(p, b_j^t)=0$.
As shown in Fig. \ref{fig:pm} (b)(c),
when pixel $p$ is enclosed by proposal $b_j^t$,
the proposal's score $F(b_j^t)$ is added to this pixel.
Finally, each pixel in this map $P$ indicates the probability to be obstacle.
By setting a parameter to segment the map into a mask,
the obstacle is eventually located by the segmented mask.
Since the low-score proposal has lower weight,
the proposed scheme avoids an extremely high response in parts of obstacles caused by a large number of proposals with similar locations,
as shown in Fig. \ref{fig:pm} (c).

\section{Evaluation}
\label{sec:eval}
\subsection{Obstacle on Reflective Ground (ORG) Dataset}
To evaluate the proposed method,
a novel dataset for Obstacle on Reflective Ground (ORG) is proposed,
which consists of 15 different indoor scenes and 42 types of obstacles,
as shown in Fig.\ref{fig:exam}.
The scenes of this dataset are mainly divided into three types:
high contrast floor,
smooth floor,
and blurry floor.
In these scenes,
the ground of each scene has varying UOs to evaluate the robustness of the algorithm.
Even,
low illumination, different patterned floor, and small obstacle also appear in this dataset.
In terms of the obstacle,
the obstacles with different sizes and materials are contained in the dataset,
which have low height and therefore fail to be perceived by 2D LiDAR.

The ORG dataset contains 223 monocular video sequences in total.
In each sequence,
several additional information is provided,
i.e.,
a set of pose information given by the wheel odometer,
the pixel-level annotation of the ground and the obstacles. 
Each video contains about 10 to 20 frames annotated,
and each object has a different category ID.
In terms of the dataset structure,
the dataset is split into a Train set (117 videos/1711 images) and a Test set (106 videos/1709 images),
each of which contains completely different scenes,
As the platform shown in Fig. \ref{fig:exam},
the monocular camera is fixed with a height of 60 \textit{cm} from the ground
and features a focal length of 2.8 mm,
a spatial resolution of 1920$\times$1080,
and a pixel size of 8 bit.
The Kobuki robot provides the wheel odometer for our method.

\begin{figure}
\centering
\includegraphics[width=0.9\linewidth]{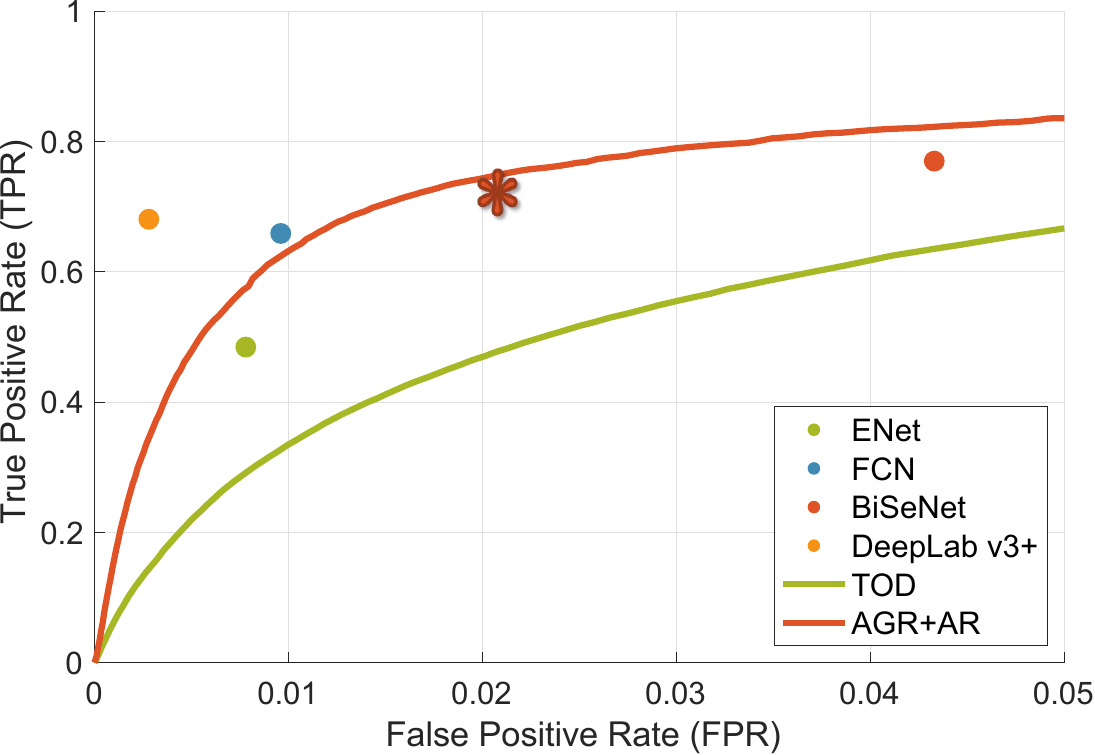}
\caption{
The pixel-level ROC of different methods.
The performance when FPR is 2\% is used to compare with other methods in the instance-level performance comparison,
which is indicated by $\star$.}
\label{fig:roc}
\end{figure}

\subsection{Metrics}
To evaluate our method fairly,
a pixel-level metric and an instance-level metric are employed to quantitatively analyze the performance of all methods,
respectively.

\noindent
\textbf{Pixel-level Metric:}
Referring to other discovery methods \cite{ICRA,LAF,UON},
we employ the Pixel-level Receiver-Operator-Characteristic (ROC) curve that measures the True-Positive Rate (TPR) under different False-Positive Rate (FPR):
\begin{equation}
TPR = \frac{TP}{GT_{obs}},\quad FPR=\frac{FP}{GT_{ground}}
\end{equation}
where $TP$ denotes the number of obstacle pixels which are correctly discovered,
and $FP$ corresponds to the number of floor pixels which are predicted as obstacle.
$GT_{obs}$ is the total number of pixels inside the obstacle proposal,
and $GT_{ground}$ is the total number of pixels labeled as the floor.

\noindent
\textbf{Instance-level Metric:}
Referring to \cite{LAF,UON},
since the pixel-level metric suffers from the bias toward object instances that cover large areas in the images,
we employ the instance-level metric to evaluate our method, namely Instance-level True-Positive Rate (ITPR) and Mean Instance-level False Positives (MIFP).
Inspired by \cite{LAF,UON},
each consecutive area in the generated segmentation is an instance.
An obstacle instance is marked correctly detected if more than 50\% of the pixels of the instance is obstacle.
An instance is considered as incorrectly detected if its overlap with the floor area is larger than 50\%.
Based on these definitions,
ITPR is defined as the fraction of obstacle instances in the ground truth map which are correctly detected.
MIFP is defined as the mean incorrectly detected instances per frame.
\begin{equation}
    ITPR = \frac{iTP}{N_{obs}},\quad 
    MIFP = \frac{iFP}{N_{img}}
\end{equation}
where $iTP$ denotes the total number of correctly detected obstacles,
and $iFP$ corresponds to the total number of false-positive instances.
$N_{obs}$ is the total number of obstacle instances,
and $N_{img}$ is the number of frames in the test set.

\subsection{Experimental Setup}
Several variants are compared to illustrate the effectiveness of our method.
AGR denotes the geometry-appearance fusion regressor $F^{ag}$,
and AR denotes the appearance based regressor $F^{a}$.
AGR+AR denotes our method that uses AR to handle the proposal with fast moving.
Note that the tree numbers $K^{a}$ and $K^{ag}$ are set to 50.
In addition,
$\tau_e$ is set to 0.8,
$\tau_{gc}$ is set to 0.1,
and $\tau_{b}$ is set to 50.
When discovering the obstacles in image $I^t$,
$q$ is set to ensure that the distance the robot moves $\|\Delta C\|$ is larger than 20 \textit{cm}.

\begin{table}[!tp]
\begin{center}
\caption{Instance-level results of all methods,
The thresholds are set according to FPR of 2\%.
Bold numbers indicate the 1-st results,
and underlined numbers for the 2-nd results.}
\begin{tabular}{@{}lcccc@{}}
\toprule
\specialrule{0em}{2pt}{2pt}
\multirow{2}{*}{Method} & ITPR$\uparrow$ /\% & MIFP$\downarrow$ & TPR$\uparrow$ /\% & FPR$\downarrow$ /\% \\
 & (Instance) & (Instance) & (Pixel) & (Pixel)\\
\specialrule{0em}{2pt}{2pt}
\midrule
\specialrule{0em}{1pt}{1pt}
FCN \cite{FCN} & 54.29 & \textbf{1.15} & 48.43 & \underline{0.78} \\
\specialrule{0em}{1pt}{1pt}
ENet \cite{ENet} & 51.43 & 55.22 & 64.88 & 0.96 \\
\specialrule{0em}{1pt}{1pt}
BiSeNet \cite{BiSeNet} & \underline{64.85} & 5.37 & \textbf{76.98} & 4.33\\
\specialrule{0em}{1pt}{1pt}
DeepLab v3+ \cite{DeepLab}& 52.02&2.48&68.05&\textbf{0.28} \\
\specialrule{0em}{1pt}{1pt}
\midrule
\specialrule{0em}{1pt}{1pt}
TOD \cite{ICRA}    & 61.72& 3.46& 47.00 & 2.01\\
\specialrule{0em}{1pt}{1pt}
AGR+AR& \textbf{77.46} & \underline{1.87} & \underline{74.23} & 1.98 \\
\specialrule{0em}{1pt}{1pt} \bottomrule
\end{tabular}
\label{table:Inst}
\end{center}
\vspace{-10pt}
\end{table}

\begin{figure*}
	\centering
	\includegraphics[width=1\linewidth]{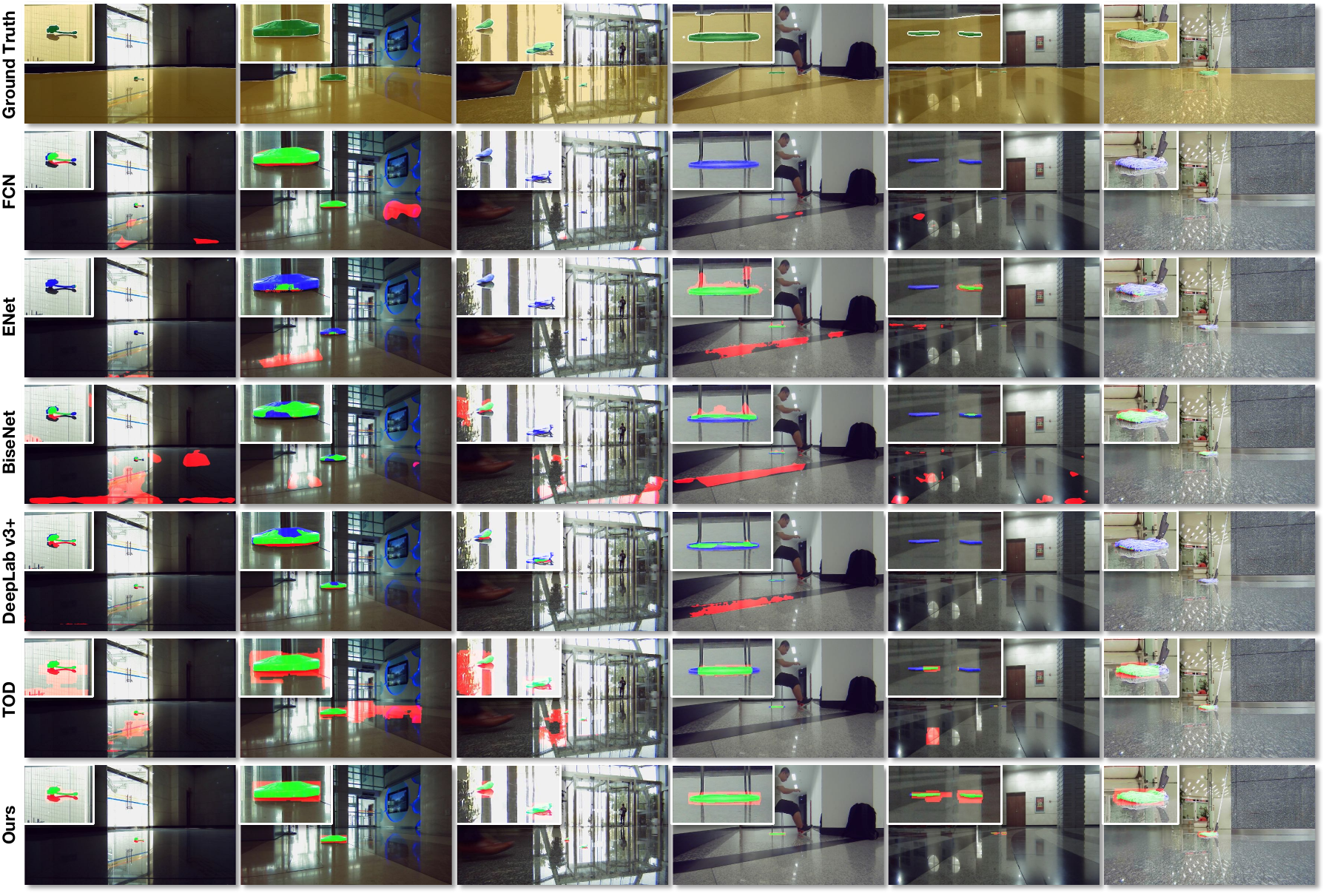}
		\caption{
		Qualitative results for obstacle discovery in the indoor environment.
		In the ground truth,
		the floor is marked in yellow,
		and green for the obstacle.
		The images could be zoomed in to clearly observe these tiny obstacles.
		In the results,
		the true-positive pixel of obstacles are marked in green,
		blue for false-negative pixels,
		and red for false-positive pixels.}
	\label{fig:final}
\end{figure*}

Furthermore,
due to lacking the source code of obstacle discovery methods \cite{LAF,UON,MergeNet},
we cannot compare our method with them.
To make a full comparison with the state-of-the-art methods,
we are inspired by the existing CNN-based obstacle discovery methods \cite{UON,MergeNet,Hua_ICCVW,JMOD2,2020LiDAR},
and employ several classic semantic segmentation networks to discover obstacles,
namely FCN \cite{FCN}, ENet \cite{ENet}, BiSeNet \cite{BiSeNet}, and Deeplab v3+ \cite{DeepLab}.
For these networks,
the parameters of the optimizer are set to the same as those proposed in their papers.
Note that ENet uses Adam \cite{adam} as the optimizer,
while other methods take stochastic gradient descent (SGD) \cite{alexnet}.
To adapt these networks to our hardware environment,
we conduct several modifications when training these networks.
For clarity,
ResNet-50 \cite{He_2016_CVPR} is employed as the backbone for Deeplab v3+.
The batch sizes of FCN and Deeplab v3+ are set to 6 and 4, respectively.
During training,
we randomly crop the image into a fixed size as input.
Specifically, the resolution for training FCN or ENet is 960$\times$540,
a half resolution of the original image,
while BiSeNet and Deeplab v3+ take the crop size 1024$\times$1024 and 513$\times$513 respectively.
For all the methods,
we use full resolution image as input during testing.
All networks are trained on two NVIDIA GeForce GTX 1080 Ti GPUs.
In addition,
the base method, namely segmenting by detection framework \cite{ICRA}, is denoted as TOD in the experiment.

\subsection{Result}
\subsubsection{Quantitative Result}
Fig.\ref{fig:roc} depicts the pixel-level ROC of different methods.
One can see that AGR+AR outperforms TOD by a large margin.
Especially,
by jointly utilizing AGR and AR,
the FPR of our method is 0.85\% when the TPR is 60\%,
a drop of about 76.94\% (from 3.69\% to 0.85\%) compared to TOD,
which proves the effectiveness of our method in avoiding the mis-detection of the UOs.
Besides,
our method even outperforms ENet \cite{ENet} and BiSeNet \cite{BiSeNet} and achieves a comparative pixel-level performance to DeepLab v3+ \cite{DeepLab} and FCN \cite{FCN}.
Note that,
since the proposed method segments obstacles by the detection,
most of the mis-detected pixels are in a rectangular area that includes the irregular-shape obstacle.
More details can be found in the instance-level result and the qualitative results.

Table \ref{table:Inst} shows the instance-level results of different methods.
For TOD and our method,
the ITPR, MIFP, and TPR are determined by setting the FPR to 2\%.
Observably,
our method achieves 15.74\% ITPR improvement over TOD and obtains 1.59 fewer false detection instances per picture than TOD.
Besides,
ENet \cite{ENet} performs poor at TPR, ITPR, and MIFP.
BiSeNet \cite{BiSeNet} obtains high TPR and ITPR at the cost of poor FPR and MIFP.
The reason is that the lightweight spatial branch of BiSeNet \cite{BiSeNet} fails to suppress the high-frequency visual information and instead retains it as noise, 
while the context branch is unable to eliminate the noise when fusing two branches' features.
As a result, more false positives are generated in high-contrast areas,
as observed in Fig. \ref{fig:final}.
On the contrary,
DeepLab v3+\cite{DeepLab} and FCN \cite{FCN} achieve better FPR and MIFP but poor TPR and ITPR.
Compared to these CNN-based methods,
although the FPR of our method is not the lowest one,
our method achieves the best performance of ITPR,
and the second-best performance of MIFP.
Thus, our method achieves a better trade-off.
Besides, this comparison also indicates that most of the false-positive pixels of our model is adjacent to the pixels of obstacle,
not scattered to the area of UOs,
which is consistent with the visualization in Fig. \ref{fig:final}.
Thus, our method avoids the wrong detection of UOs better than all others.
\begin{figure}
	\centering
	\includegraphics[width=0.9\linewidth]{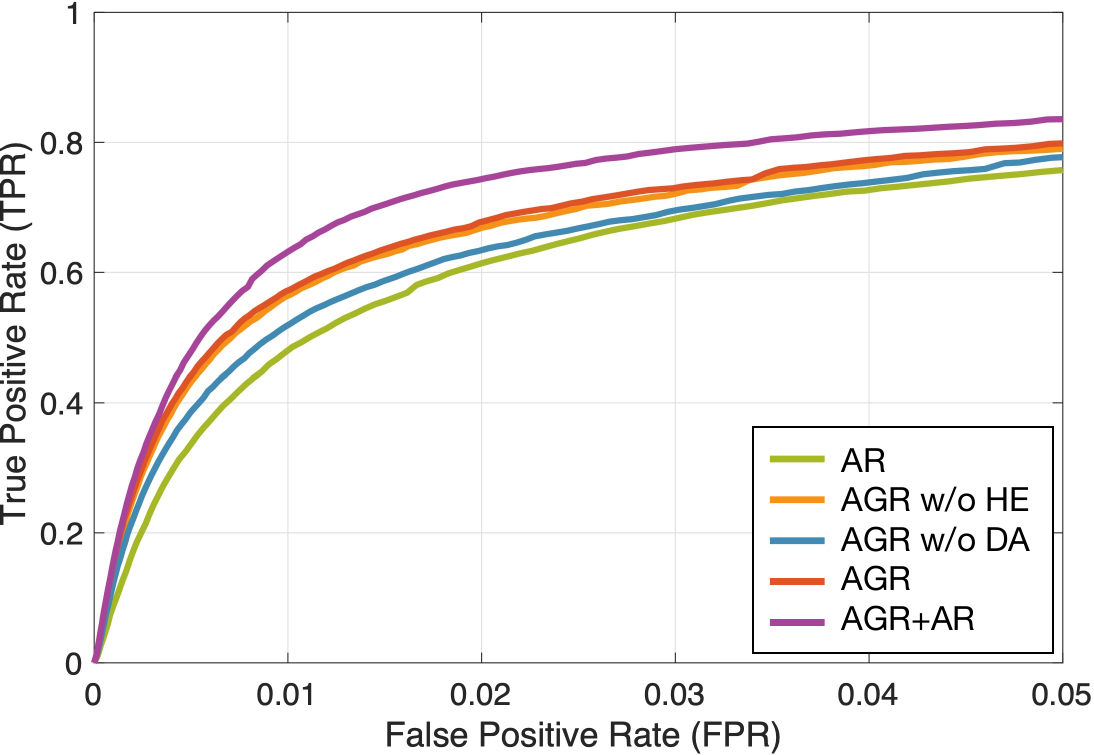}
		\caption{The pixel-level ROC of models using different features. 
		}
	\label{fig:roc_abla}
\end{figure}

\begin{table}[!tp]
\begin{center}
\caption{Instance-level results of models using different features.
Bold numbers indicate the $1^{st}$ results,
and underlined numbers for the $2^{nd}$ results.
w/o means 'without',
HE for homography error,
DA for deviation angle.}
\begin{tabular}{@{}lcccc@{}}
\toprule
\specialrule{0em}{2pt}{2pt}
\multirow{2}{*}{Method} & ITPR$\uparrow$ /\% & MIFP$\downarrow$ & TPR$\uparrow$ /\% & FPR$\downarrow$ /\% \\
 & (Instance) & (Instance) & (Pixel) & (Pixel)\\
\specialrule{0em}{2pt}{2pt}
\midrule
\specialrule{0em}{1pt}{1pt}
AR & 64.47 & \textbf{1.86} & 61.39 & 2.00 \\
\specialrule{0em}{1pt}{1pt}
AGR w/o HE & 76.46 & 2.07 & 66.92 & 2.00\\
\specialrule{0em}{1pt}{1pt}
AGR w/o DA & 73.24 & 1.91 & 63.28 & 2.00\\
\specialrule{0em}{1pt}{1pt}
AGR & \underline{76.83}& 1.94& \underline{67.67}& \underline{1.99} \\
\specialrule{0em}{1pt}{1pt}
AGR+AR& \textbf{77.46} & \underline{1.87} & \textbf{74.23} & \textbf{1.98} \\
\specialrule{0em}{1pt}{1pt} \bottomrule
\end{tabular}
\label{table:Inst_abla}
\end{center}
\end{table}

\subsubsection{Qualitative Result}
Fig.\ref{fig:final} depicts the qualitative result of different methods in indoor scenarios with different UO in the Test subset.
Note that,
we take the obstacle segmentation of TOD and our method by fixing the FPR to 2\%,
which is consistent with the setting of Table \ref{table:Inst}.
In the visualization of results,
the green pixels denote the true-positive pixels of obstacle,
the red pixels for the false-positive pixels,
and the blue pixels for the false-negative pixels.
Besides,
in the ground truth,
the green pixels and the yellow pixels indicate obstacle and floor, respectively.
The area near the obstacle is zoomed in for clear observation.

\begin{figure}
	\centering
	\includegraphics[width=1\linewidth]{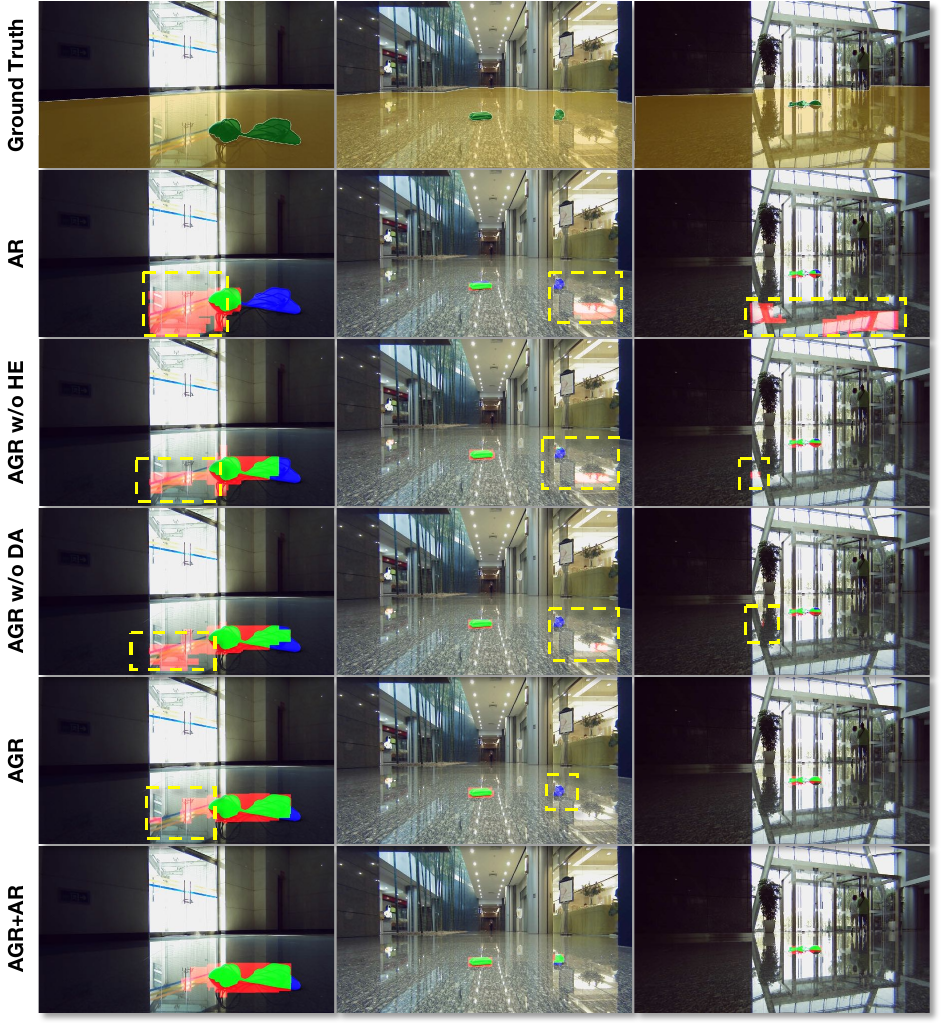}
		\caption{The visualized result of different variants.
		In the ground truth,
		the floor is marked in yellow,
		and green for the obstacle.
		In the results,
		the true-positive pixel of obstacles are marked in green,
		blue for false-negative pixels,
		and red for false-positive pixels.
		The yellow dotted boxes enclose the wrong prediction.}
	\label{fig:vis_abla}
\end{figure}

The first column shows a bunch of data lines on a smooth floor,
where the reflection causes an extremely large color difference.
It can be seen that all other methods are confused by the reflective door.
The CNN-based segmentation methods suffer from a lot of false-positive pixels,
and TOD is also degraded due to the wrong segmentation of the reflective door handle.
Although our method generates a few false positives nearby the obstacle,
no false positive exists in the area of UOs.
The reason is that the proposed features well represent the obstacle and the reflection.
The second column shows a guardrail base in the lobby.
The floor reflects some fuzzy highlights due to its matte surface.
FCN wrongly takes the fake TV as the obstacle.
ENet, BiSeNet, and TOD not only generate many false-positive pixels,
but also fail to segment the real obstacle.
In contrast,
our method corrects the misclassification of TV and highlight reflection,
and segments the complete obstacle.
The third column depicts a tube of hand cream and a pair of glasses on the floor of the office building.
Due to the extremely smooth surface,
the floor reflects all the scenes in the real world,
making it hard to distinguish the real obstacles.
FCN and ENet fail to discover the two obstacles.
BiSeNet discovers one of the obstacles,
but suffer from the wrong classified pixels inside the area of reflection.
DeepLab v3+ successfully discovers them all,
but misses several obstacle pixels.
TOD is unable to distinguish the reflection and the real obstacle.
Compared to them,
our method segments all the obstacles and avoids false-positive pixels in the area of reflection.
The fourth column shows a round stool on the ground with multiple textures.
The four thin legs and the base of this stool can hardly be captured by the LiDAR of the robot.
The reflection is not obvious,
but the different texture obviously confuses all the CNN-based methods.
TOD performs poorly due to the inaccurate segmentation of the obstacle,
but our method avoids all the problems.
The fifth column shows two cellphones.
The floor is full of reflections and multiple textures.
And the dim illumination makes it harder to recognize the cellphones on the ground.
Intuitively,
FCN, ENet, BiSeNet, and TOD cannot discover all the cellphones,
while failing to avoid the false positives.
DeepLab v3+ avoids the wrong classification to the obstacle pixel,
but is unable to discover the two obstacles.
By contrast,
our method avoids the false positives,
and discovers the two cellphones.
The last column shows a mop leaning against the wall.
Due to the similar appearance with the ground,
almost all the methods cannot discover it totally.
However, our method can capture this mop completely.

\begin{table}[!tp]
\begin{center}
\caption{Instance-level results when using various $\tau_b$ in generating probability map.
The thresholds are set according to FPR of 2\%.
Bold numbers indicate the 1-st results,
and underlined numbers for the 2-nd results.
}
\begin{tabular}{@{}lcccc@{}}
\toprule
\specialrule{0em}{2pt}{2pt}
\multirow{2}{*}{Method} & ITPR$\uparrow$ /\% & MIFP$\downarrow$ & TPR$\uparrow$ /\% & FPR$\downarrow$ /\% \\
 & (Instance) & (Instance) & (Pixel) & (Pixel)\\
\specialrule{0em}{2pt}{2pt}
\midrule
\specialrule{0em}{1pt}{1pt}
Original \cite{ICRA}  & 76.23& 2.46 & 65.52& 2.00 \\
\specialrule{0em}{1pt}{1pt}
\midrule
\specialrule{0em}{1pt}{1pt}
Top 10  & 71.47& \textbf{1.72} & 70.30& 2.00 \\
\specialrule{0em}{1pt}{1pt}
Top 50 & 77.46 & \underline{1.87} & \underline{74.23} & \textbf{1.98} \\
\specialrule{0em}{1pt}{1pt}
Top 100 & \textbf{78.37}& 1.97 & \textbf{74.36}& \underline{1.99} \\
\specialrule{0em}{1pt}{1pt}
Top 200 & \underline{78.00}& 2.05 & 73.10& \underline{1.99} \\
\specialrule{0em}{1pt}{1pt}
Top 500 & 76.55& 2.19 & 71.35& 2.02 \\
\specialrule{0em}{1pt}{1pt}
\bottomrule
\end{tabular}
\label{table:Inst_topN}
\end{center}
\end{table}

In general,
it can be seen from all the above results that our method successfully avoids the confusion brought by the ground reflection,
and discovers the obstacles better than other methods.
It is noteworthy that our method obtains a few false-positive obstacle pixels in the rectangle region enclosing the obstacle.
The reason is that we segment the obstacle by detection,
the region proposal inevitably covers these false-positive pixels.
Fortunately,
due to connection with the obstacles,
these few false-positive pixels do not affect the robot navigation at all.
Moreover,
this observation also exposits the highest ITPR and low MIFP mentioned in the last subsection.

\begin{figure}
\centering
\includegraphics[width=0.9\linewidth]{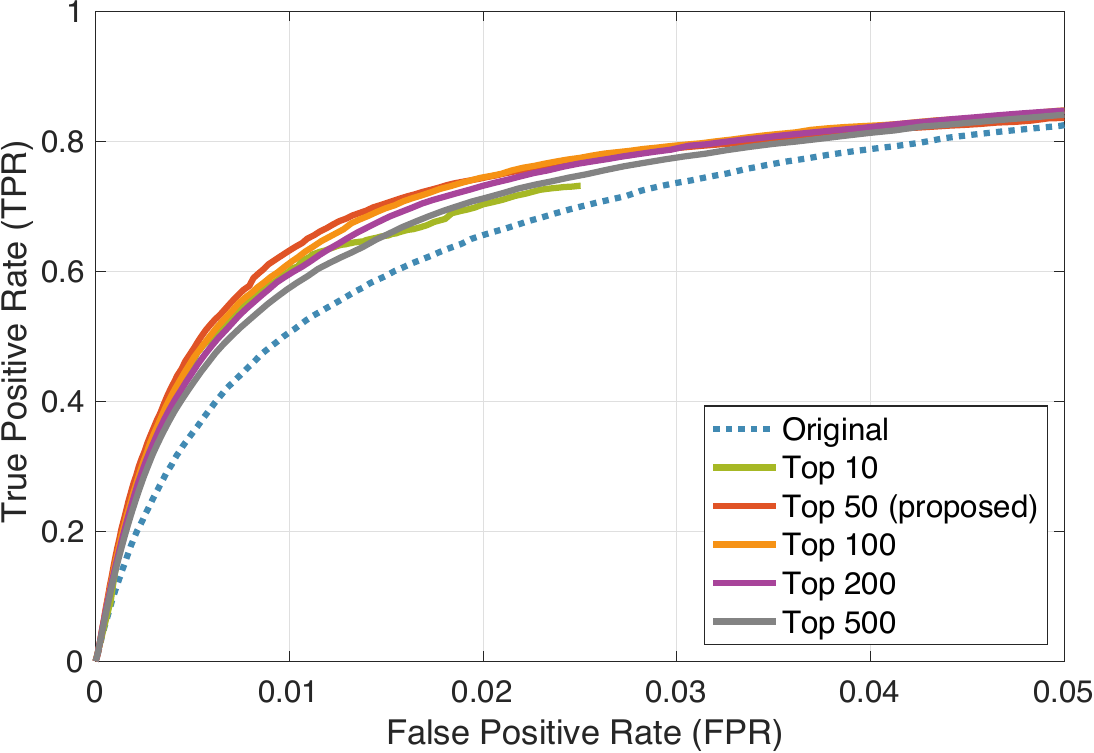}
\caption{The pixel-level ROC of variants with different numbers of top proposals in generating the probability map.}
\label{fig:roc_abla3}
\end{figure}

\subsection{Analysis and Discussions}
\subsubsection{Effectiveness of appearance-geometry feature}
\label{sec:ablafeat}
In this paper,
we propose a unified appearance-geometry feature representation for bounding box proposals.
The effectiveness of the proposed feature is evaluated in the comparisons in Fig. \ref{fig:roc_abla} and Table \ref{table:Inst_abla}.
Intuitively,
AR only employs the appearance features,
and suffers from the low ITPR and TPR.
The reason is that the lack of geometry feature leads to a reduced ability to distinguish obstacles from UOs.
By additionally using the homography error (HE),
the TPR is further improved by 1.89\%,
and 8.77\% for ITPR.
By additionally using the deviation angle (DA),
the TPR is further improved by 5.53\%,
and 11.99\% for ITPR.
The two results demonstrate the effectiveness of homography error and deviation angle, respectively.
By jointly using the two features,
the improvement of TPR is 6.28\% (comparing AGR and AR),
and that of ITPR for 12.36\%.
Besides,
the model using AGR and AR,
namely the proposed model,
achieves the best performance,
6.56\% TPR higher than the second-best result,
and 0.63\% ITPR higher than the second best result.
The reason is that,
by using AR to avoid the misclassification of proposals with low confidence,
the proposed model better utilizes the appearance-geometry feature representation.

Fig. \ref{fig:vis_abla} shows the visualized results of the mentioned variants.
The first column shows a wired mouse on the reflective ground with extreme illumination change.
Due to the insufficient expression of feature to reflection,
AR, AGR w/o DA, and AGR w/o HE wrongly segment the reflection.
AGR performs better than the prior variants,
and AGR+AR has the best result.
The second column shows a watch and a wallet on the floor of the mall.
The size of the watch is small,
which makes almost all the variants fail to capture it.
AGR+AR not only segments the watch,
but also avoids the false positives on the floor.
The third column depicts a mouse on the ground with complex reflection.
AR is completely incapable of avoiding false detection of reflection.
\xf{AGRs} with only HE or DA also cannot completely avoid false detection.
AGR and AGR+AR perform well in this scene.

\begin{figure}
\centering
\includegraphics[width=1\linewidth]{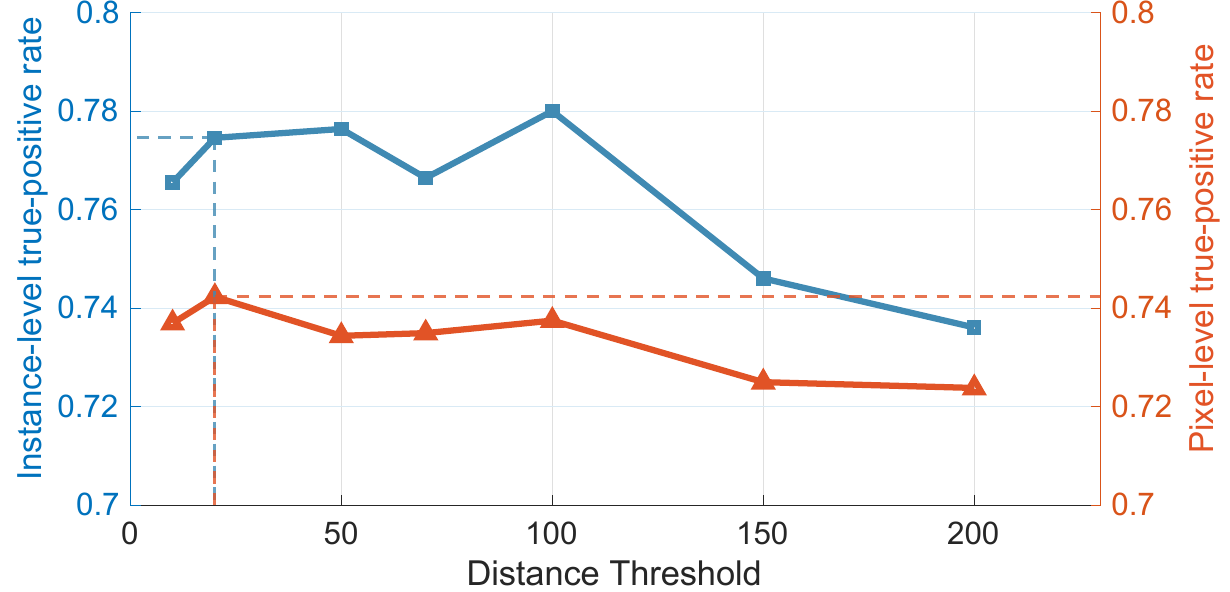}
\caption{The pixel-level and instance-level results of variants with different distance threshold when determining $q$. We select the results when FPR is fixed to 2\%.}
\label{fig:roc_dist_abla}
\end{figure}

\subsubsection{Different settings of weighted-decay probability map}
In this paper,
the weighted-decay probability generation scheme is proposed to avoid that the final segmentation concentrates too much on parts of obstacles.
To validate its effectiveness,
the original generation scheme in \cite{ICRA,tip} is employed to compare with the proposed method.
It firstly uses NMS to remove redundant proposals,
then accumulates the top 50\% proposals to generate the probability map.
Besides,
we take five number of top proposals in the generation process to find the best parameter.

Fig. \ref{fig:roc_abla3} and Table \ref{table:Inst_topN} show the results when using different parameters in generating the probability map.
Firstly,
by comparing the original generation method and the proposed one,
it can be seen that our method fully outperforms the original generation method.
Secondly,
as the number of top proposals increases,
the pixel-level accuracy first improves and then decreases,
and the instance-level accuracy follows the same way.
Intuitively,
top 50, top 100, and top 200 achieve better performances than others.
Eventually,
we consider that the optimal number of top proposals is 50,
which is brought by two reasons:
\begin{itemize}
    \item It performs best when FPR ranges from 0\% to 2\%,
    which makes our model obtains fewer wrong segmentation pixels.
    \item Although its instance-level accuracy is slightly poorer than top 100 and top 200,
    it obtains fewer false-positive instances.
\end{itemize}

\begin{figure}
\centering
\includegraphics[width=1\linewidth]{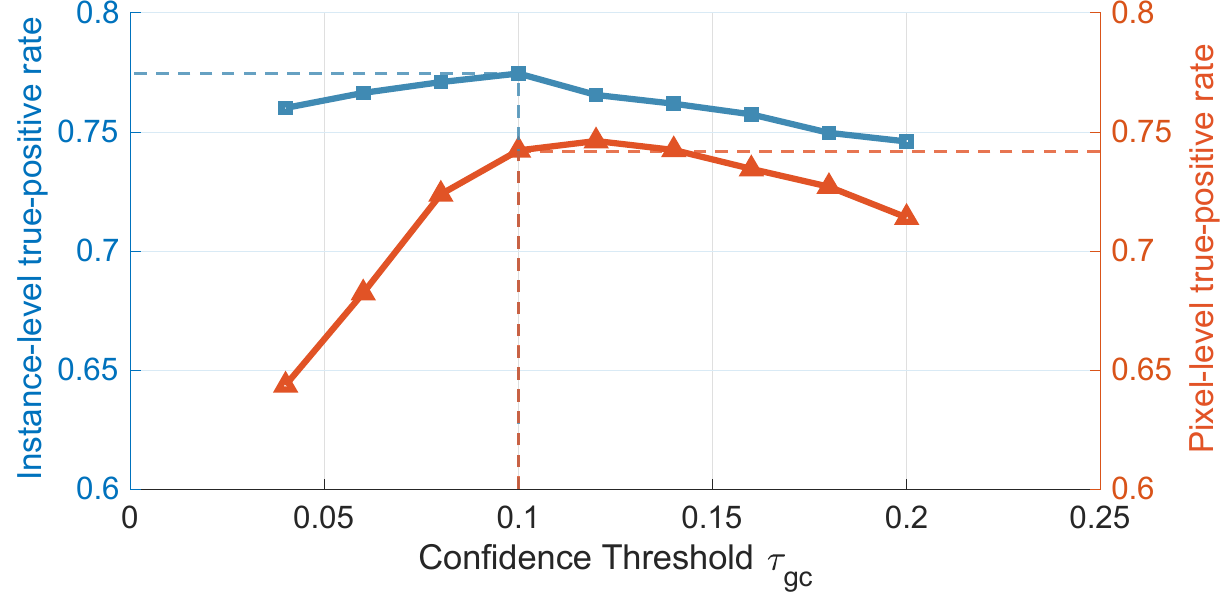}
\caption{The pixel-level and instance-level results of variants with different $\tau_{gc}$. We select the results when FPR is fixed to 2\%.}
\label{fig:roc_tau_abla}
\end{figure}

\begin{figure*}
\centering
\includegraphics[width=0.95\linewidth]{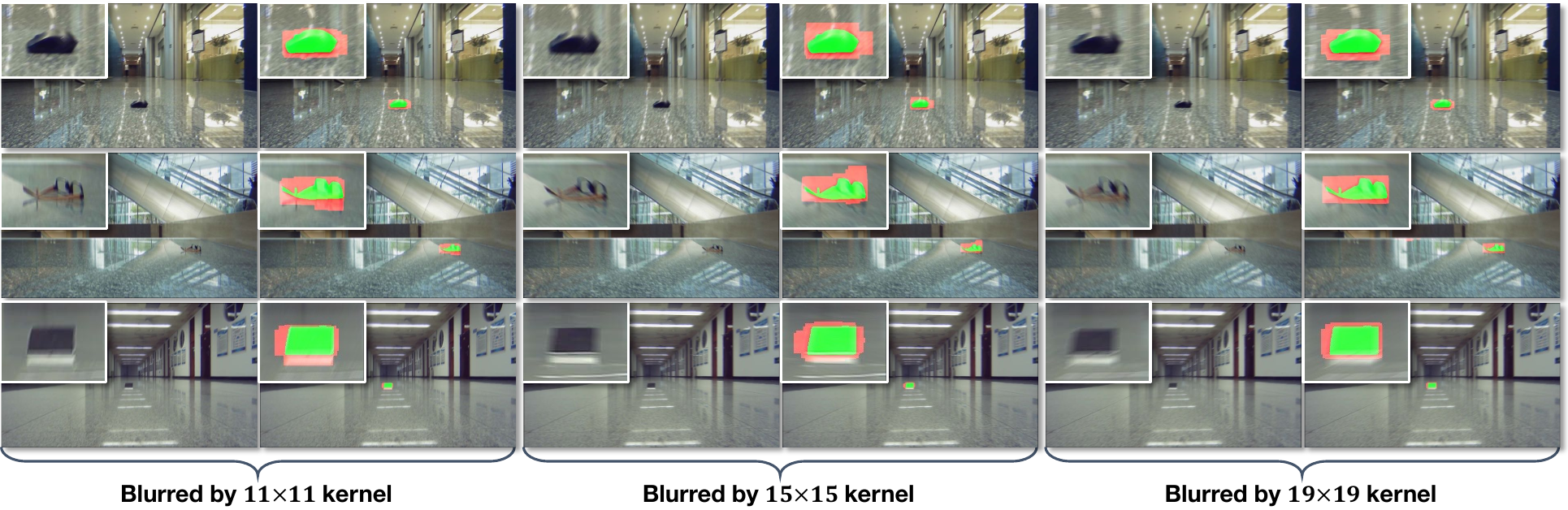}
\caption{The visualized results of variants adding different levels of the motion blur.
In the results,
the true-positive pixel of obstacles are marked in green,
blue for false-negative pixels,
and red for false-positive pixels.}
\label{fig:vis_motionblur}
\end{figure*}

\subsubsection{Different settings of distance threshold}
In our method, the parameter $q$, which represents the frame interval of two consecutive images,
is set to ensure that the distance the robot moves is larger than a certain distance threshold.
Fig. \ref{fig:roc_dist_abla} displays the pixel-level and instance-level results of the model with different distance thresholds.
The results illustrate that the TPR and ITPR decrease significantly when the threshold is lower than 20 or larger than 100.
The reason is that excessive movement would result in a large change in the camera view.
Furthermore, there are no significant changes in accuracy when the threshold ranges from 20 to 100,
and the model that uses a threshold of 20 achieves the best TPR and the third best ITPR. 

\subsubsection{Different settings of confidence threshold $\tau_{gc}$}
The variable $\tau_{gc}$ represents the confidence threshold used to select between AR or AGR for scoring bounding boxes.
Fig. \ref{fig:roc_tau_abla} illustrates the pixel-level ROC curve and instance-level results obtained by varying $\tau_{gc}$ from 0.04 to 0.20.
Notably, setting $\tau_{gc}$ to 0.1 yields the highest ITPR, lowest MIFP, and third-highest TPR.
Furthermore, when $\tau_{gc}$ is set below 0.08,
TPR and ITPR decrease significantly due to insufficient usage of AGR,
and when $\tau_{gc}$ exceeds 0.12,
TPR and ITPR continue to decrease due to the limited representation ability of the low-confidence geometry feature.

\begin{figure}
\centering
\includegraphics[width=1\linewidth]{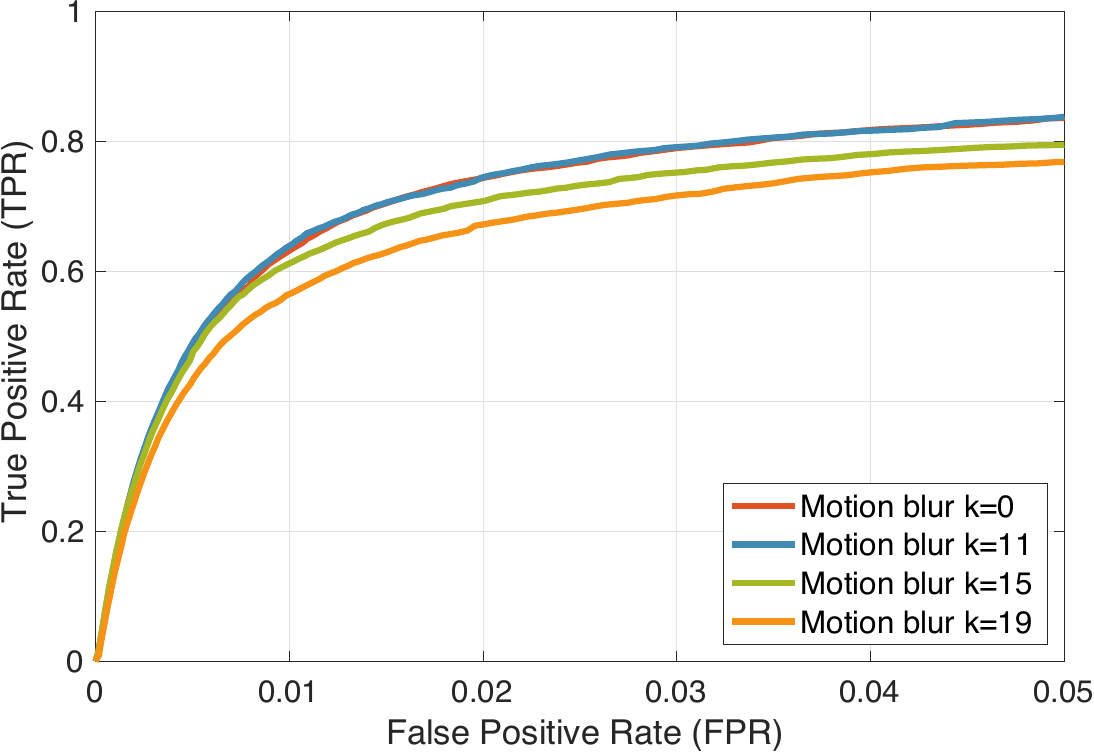}
\caption{The pixel-level ROC of variants impacted by different level of motion blur.}
\label{fig:roc_motionnoise}
\end{figure}

\subsubsection{Robustness to motion blur noise}
In actual application,
when the robot moves on the ground,
the mounted camera is inevitably impacted by the fast motion and the violent shaking.
In these cases,
the long exposure time of the camera causes the blurred image,
which presents the obstacle discovery method with challenges.
To verify the robustness of our method to the motion blur,
the random motion blurring scheme \cite{Hua_ICCVW} is employed to introduce the noise in the testing phase,
which uses the point spread function \cite{Jia_2007} to simulate the motion blur of image.
More specifically,
we utilize a blurring filter with $\mathbf{k}\times\mathbf{k}$ size and $\mathbf{\theta^*}$ direction to blur the image before being fed into our method,
where $\mathbf{k}$ determines the intensity of motion blur.
% and $\theta^*$ denotes the direction of motion.
According to the resolution of the proposed dataset and our experience,
most of the blur kernels are within 15 pixels in size.
Thus,
the kernel size used for the testing is 11, 15, and 19 pixels,
and the blur direction ranges from $0^{\circ}$ to $\pm180^{\circ}$ randomly.
Note that,
our model is trained on the image without blurring.

\begin{table}[!tp]
\begin{center}
\caption{Instance-level result of models impacted by different levels of motion blur.
The thresholds are set according to FPR of 2\%.
Bold numbers indicate the 1-st results,
and underlined numbers for the 2-nd results.
}
\begin{tabular}{@{}lcccc@{}}
\toprule
\specialrule{0em}{2pt}{2pt}
\multirow{2}{*}{\tabincell{l}{Kernel size for\\motion blur}} & ITPR$\uparrow$ /\% & MIFP$\downarrow$ & TPR$\uparrow$ /\% & FPR$\downarrow$ /\% \\
 & (Instance) & (Instance) & (Pixel) & (Pixel)\\
\specialrule{0em}{2pt}{2pt}
\midrule
\specialrule{0em}{1pt}{1pt}
$\mathbf{k}=0$  & \textbf{77.46} & \underline{1.87} & 74.23 & \textbf{1.98} \\
\specialrule{0em}{1pt}{1pt}
$\mathbf{k}=11$ & \underline{68.84} & 1.97 & \underline{74.37} & \underline{1.99} \\
\specialrule{0em}{1pt}{1pt}
$\mathbf{k}=15$ & 64.04 & 1.97 & 70.81& 2.01 \\
\specialrule{0em}{1pt}{1pt}
$\mathbf{k}=19$ & 57.32 & \textbf{1.85} & 67.21 & 2.00 \\
\specialrule{0em}{1pt}{1pt}
\bottomrule
\end{tabular}
\label{table:Inst_motionnoise}
\end{center}
\end{table}

\begin{table*}[!tp]
\begin{center}
\caption{Instance-level results when adding noise to the robot motion.
The thresholds are set according to FPR of 2\%.}
\begin{tabular}{@{}ccccccc@{}}
\toprule
\specialrule{0em}{2pt}{2pt}
\multirow{2}{*}{Variants}&\multirow{2}{*}{Translation Noise} & \multirow{2}{*}{Rotation Noise}& ITPR$\uparrow$ /\% & MIFP$\downarrow$ & TPR$\uparrow$ /\% & FPR$\downarrow$ /\% \\
&& & (Instance) & (Instance) & (Pixel) & (Pixel)\\
\specialrule{0em}{2pt}{2pt}
\midrule
\specialrule{0em}{1pt}{1pt}
1 & - & - & 77.46& 1.87 & 74.23 & 1.98 \\
\specialrule{0em}{1pt}{1pt}
2 & $[-5\%,0]\cup[0,5\%]$ & - & 77.69 & 1.89 & 74.44 & 2.02 \\
\specialrule{0em}{1pt}{1pt}
3 & $[-10\%,-5\%]\cup[5\%,10\%]$ & - & 77.01 & 1.88 & 74.06 & 1.99 \\
\specialrule{0em}{1pt}{1pt}
4 & - & $[-0.007,-0.003]\cup[0.003,0.007]$ & 77.51 & 1.87 & 74.17 & 2.01 \\
\specialrule{0em}{1pt}{1pt}
5 & - & $[-0.011,-0.007]\cup[0.007,0.011]$ & 77.46 & 1.86 & 74.26 & 2.00 \\
\specialrule{0em}{1pt}{1pt}
6 & - & $[-0.016,-0.011]\cup[0.011,0.016]$ & 77.10 & 1.85 & 73.42 & 2.00 \\
\specialrule{0em}{1pt}{1pt}
7 &$[-5\%,0]\cup[0,5\%]$&$[-0.011,-0.007]\cup[0.007,0.011]$& 77.41 & 1.87 & 73.99 & 2.00 \\
\specialrule{0em}{1pt}{1pt}
8 &$[-10\%,-5\%]\cup[5\%,10\%]$&$[-0.016,-0.011]\cup[0.011,0.016]$& 76.64 & 1.85 & 73.43 & 2.01 \\
\specialrule{0em}{1pt}{1pt}
\bottomrule
\end{tabular}
\label{table:Inst_odomnoise}
\end{center}
\end{table*}

Fig. \ref{fig:roc_motionnoise} and Table \ref{table:Inst_motionnoise} illustrate the result of our method affected by the motion blur.
Intuitively,
when the size of the blurring filter is $11\times11$,
the pixel-level accuracy is barely affected,
and the instance-level accuracy is reduced by 8.62\%.
The reason is that several small obstacles are smaller than the blurring kernel,
and thus their visual information is lost.
The failure to discover these extremely tiny obstacles has a small impact on pixel-level accuracy,
but has a great impact on instance-level accuracy.
When the size of the blurring filter is enlarged to $15\times15$,
both the instance-level and pixel-level accuracy are further reduced by around 4\%.
When the intensity of blurring is maximized,
the accuracy is further reduced.
The above comparison proves that although our method suffers from the noise given by motion blur,
the accuracy is still comparable to other methods,
which proves the robustness of our method.

Fig.\ref{fig:vis_motionblur} shows three scenes blurred by different kernels.
The first row depicts a wireless mouse.
It can be seen that,
since it is close to our robot,
this mouse can be discovered well in all intensities of motion blur.
The second row shows a pair of glasses,
it is farther from the robot than the mouse in the first row.
Observably,
even if some false-positive pixels are generated when using $19\times19$ kernel,
the pair of glasses can be completely segmented.
The last row shows a pad,
which is far from the robot.
Intuitively,
the visual information is largely blurred.
Our method still successfully discover this obstacle.

\begin{figure}
\centering
\includegraphics[width=0.9\linewidth]{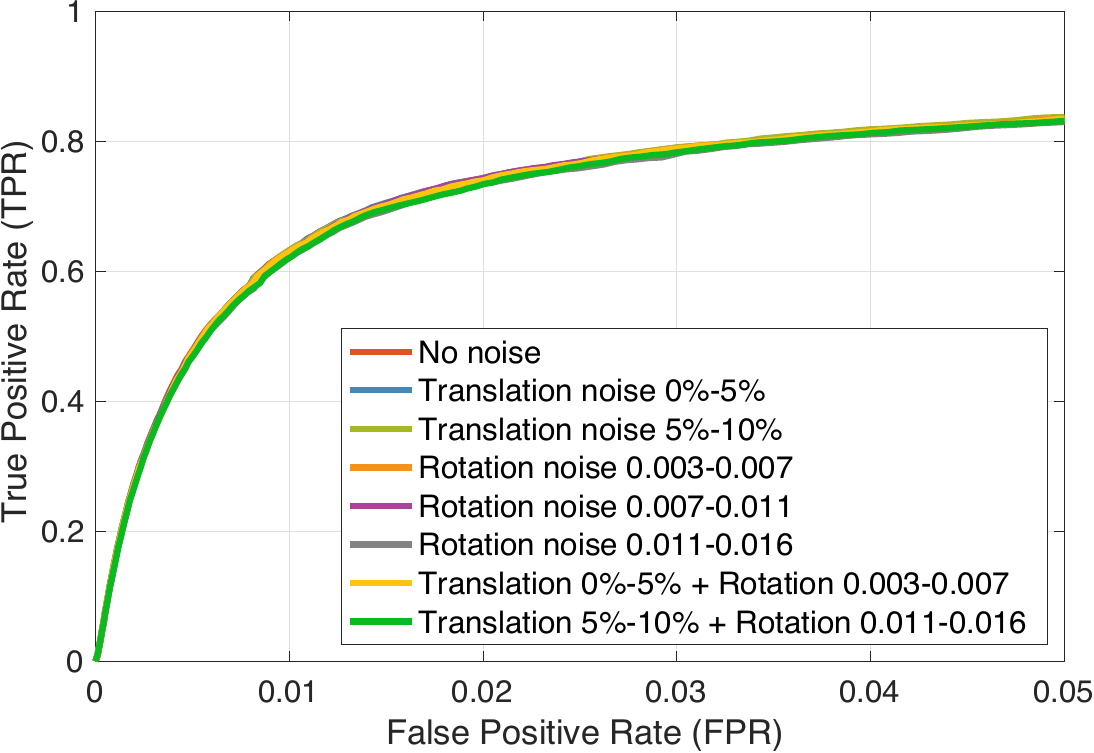}
\caption{The pixel-level ROC when adding different levels of the motion noise.}
\label{fig:roc_odomnoise}
\end{figure}

\subsubsection{Robustness to robot motion noise}
Since our method exploits the robot motion to figure the motion of the mounted camera,
theoretically,
the accuracy of the robot motion determines the performance of our method.
Hence,
we add the noise of different intensities on the robot motion to validate the robustness of our method.
In more detail,
the robot motion can be decomposed into translation and rotation.
Assuming that the time difference between the wheel odometer and the camera is less than 10 ms,
the robot moves at a speed of 200 mm per second,
and the maximum angular velocity is 90 degrees per second,
the maximum angular error caused by the unsynchronized sensor is $90^{\circ}/s\times0.01s = 0.9^{\circ} \approx 0.016 rad$.
Hence, we add the random noise of two intensities on the translation of robot,
i.e., $[-5\%,0]\cup[0,5\%]$ and $[-10\%,-5\%]\cup[5\%,10\%]$.
In addition,
we add the noise of three intensities on the rotation of robot, i.e.,
$[-0.007,-0.003]\cup[0.003,0.007]$,
$[-0.011,-0.007]\cup[0.007,0.011]$,
and $[-0.016,-0.011]\cup[0.011,0.016]$.
Finally, we add the maximum translation noise and rotation noise on the motion of the robot.

\begin{figure}
\centering
\includegraphics[width=0.9\linewidth]{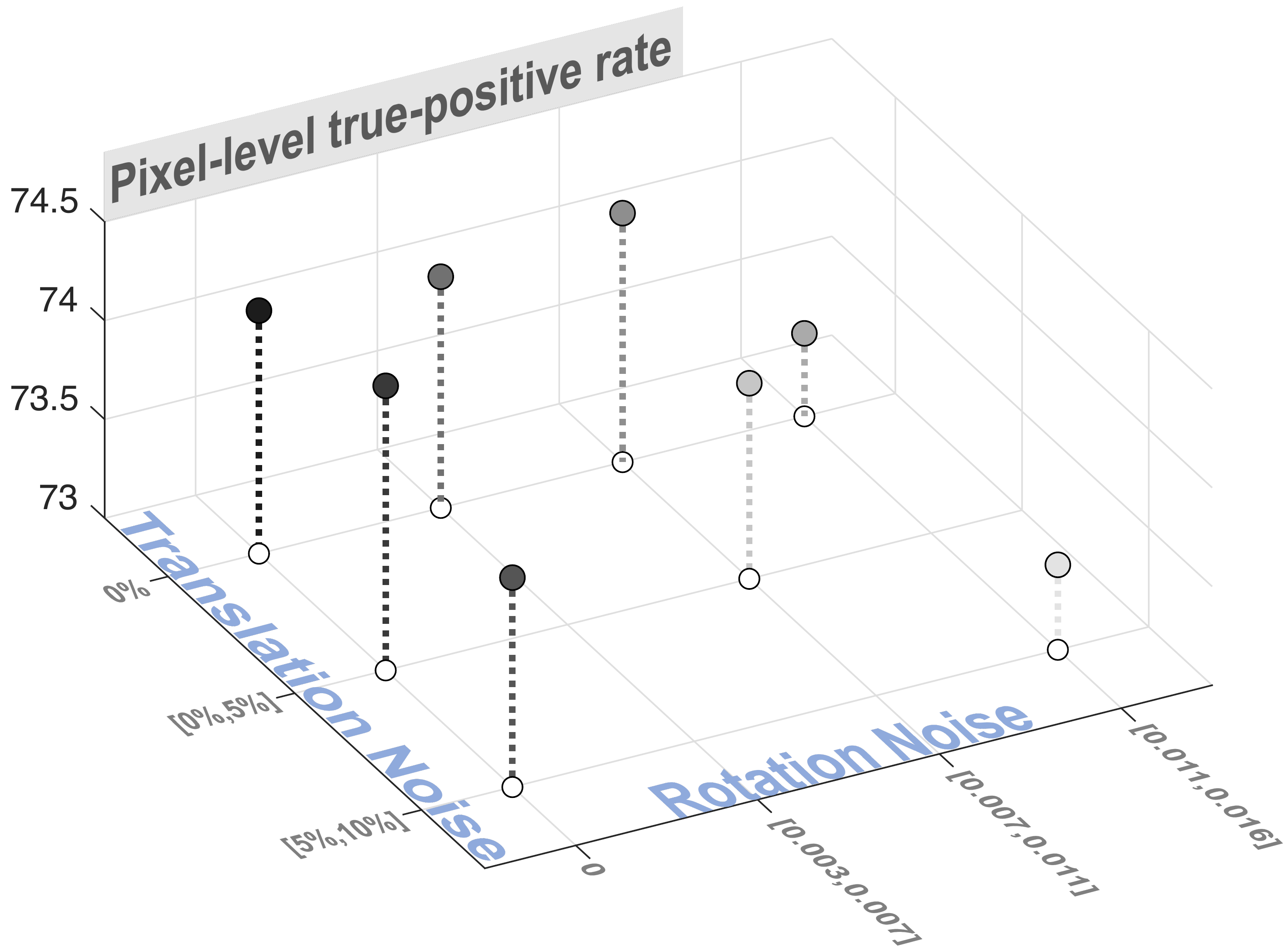}
\caption{Pixel-level true-positive rate (TPR) of models disturbed by translation and rotation noise.
The gray strength and the height of the points indicate the TPR of the variants.}
\label{fig:roc3d_motionnoise}
\end{figure}

Table \ref{table:Inst_odomnoise} and Fig.\ref{fig:roc_odomnoise} show the effect of noise on the performance,
and Fig. \ref{fig:roc3d_motionnoise} presents the TPR in a three-dimensional coordinate system.
Observably,
as the noise added to the translation increases from $0$ to $10\%$,
both TPR and ITPR decrease slightly.
Similarly, the TPR is also slightly down when rotation noise is lower than 0.011.
The performance degradation is so insignificant that our model is barely affected.
Finally, when both types of noises are increased simultaneously,
with magnitudes greater than $5\%$ and $0.011$, respectively,
both TPR and ITPR experience a more severe drop although the ITPR is only reduced by as much as $0.82\%$ and $0.81\%$ for the pixel-level accuracy.
Overall, we infer that the confidence interval for translation noise is $[-5\%, 5\%]$,
and the confidence interval for rotation noise is $[-0.011, 0.011]$.

% The instance-level accuracy is only reduced by as much as 0.82\%,
% and 0.81\% for the pixel-level accuracy.
% All these comparisons prove the robustness of our method.

\subsubsection{Inference time analysis}
Our method is implemented in MATLAB and runs on a PC with 16GB memory and an AMD Ryzen 2700 CPU.
The current implemented version of our method fails to run in real-time with an input resolution of $1920\times1080$.
But we believe that the proposed method can achieve real-time performance in the C++ implementation with parallel computing.
The inference time of each component is shown in Table \ref{table:time}.
Specifically, our method consists of five parts, edge detection, proposal extraction, feature extraction, AGFM, and obstacle occupied map generation.

The first two parts employ the original occlusion edge detection \cite{IS} and object-level proposal \cite{OLP}, which account for the majority of time consumption.
In the third part, the extraction of appearance features takes 2.339 seconds,
while the extraction of geometry features takes 1.168 seconds.
Finally, the random forest and obstacle occupied probability map generation take 0.058 seconds and 0.016 seconds, respectively.
In summary, the first three parts take up most of the overall time cost,
namely, 9.744 seconds, while the basic algorithms \cite{IS,OLP,ICRA} is the core factor.

In principle,
the proposed features can be calculated using integral images and can be computed in parallel with a time complexity of $\mathcal{O}(n)$, while part appearance features that cannot be computed using integral images have a time complexity of $\mathcal{O}(nl)$,
where $n$ denotes the number of proposal and $l$ for pixel number of a proposal.
Therefore, we infer that the reason for the current slower speed is the lack of optimized compilation processes and parallel computing.

% The first two parts and the fourth part employ the original occlusion edge detection \cite{IS}, object-level proposal \cite{OLP}, respectively.
% In the feature extraction parts, our method additionally costs 1.168 s per image.
% In the obstacle-occupied map generation parts, our method additionally spends 0.009 s per image. 

\begin{table}[!tp]
\begin{center}
\caption{Inference time of our method and the basic methods.}
\resizebox{\linewidth}{!}{
\begin{tabular}{@{}lccc@{}}
\toprule
\specialrule{0em}{2pt}{2pt}
Modules                          & Time of basic method & Basic method  & Time      \\
\specialrule{0em}{1pt}{1pt}
\hline
\specialrule{0em}{1pt}{1pt}
Edge Detection                   & 3.411 s  & \cite{IS}     & 3.411 s   \\
\specialrule{0em}{1pt}{1pt}
Proposal Extraction              & 2.826 s  & \cite{OLP}    & 2.826 s   \\
\specialrule{0em}{1pt}{1pt}
Feature Extraction               & 2.339 s  & \cite{ICRA}   & 3.507 s   \\
\specialrule{0em}{1pt}{1pt}
AGFM                             & 0.053 s  & \cite{DF}     & 0.058 s   \\
\specialrule{0em}{1pt}{1pt}
Probability Map & 0.007 s   & \cite{ICRA}                    & 0.016 s   \\
\specialrule{0em}{1pt}{1pt}
\hline
\specialrule{0em}{1pt}{1pt}
ALL                              & 8.636 s  & -             & 9.818 s   \\
\specialrule{0em}{1pt}{1pt}
\bottomrule
\end{tabular}}
\label{table:time}
\end{center}
\vspace{0pt}
\end{table}

% It can be seen that the proposed algorithm costs most of the time in the edge detection, proposal generation, and feature extraction parts.
% The feature extraction part also cost more than 3 second per image due to the lack of parallel computing.
% In future work, we will integrate a deep learning based model into feature extraction and proposal generation to expedite accelerating the overall algorithm structure as much as possible.
% We are confident that this method can achieve real-time computation in the CUDA implementation.

\section{Conclusion and Future Work}
In this paper,
a novel method is proposed to discover the obstacles on the reflective ground.
To construct the feature representation that reveals the difference between obstacles and UOs,
we first propose a ground detection scheme with pre-calibration,
and introduce the ground-pixel parallax to represent the location of an occlusion edge point relative to the ground.
Subsequently,
by aggregating the parallax and the appearance cue,
we propose a unified appearance-geometry feature representation for object bounding box proposal.
Then, an appearance-geometry fusion model is proposed to locate the obstacle,
meanwhile, avoids concentrating too much on parts of obstacles.
Finally, a novel ORG dataset is proposed to evaluate our method,
which is the first dataset focusing on the obstacle discovery on reflective ground.

This paper specializes in the problem of discovering obstacle on reflective ground.
In the future,
we are going to extend our method to address more challenges,
such as small obstacle discovery,
mirror obstacle discovery,
and dynamic obstacle discovery,
by a unified discovery framework.
In this unified framework,
monocular depth prediction \cite{Saxena2008tc,Bian2021wb,depthpr} and dynamic objects tracking \cite{icipTracker,ijcv} will be combined to further safeguard robot navigation.
Additionally, 
to accelerate the running speed,
we will adopt deep feature extraction and region proposal methods in future work, and implement parallel computation of features using C++.

\section*{Data Availability Statement}
The code of the MATLAB implementation and datasets generated during the current study are available in the GitHub repository,
\url{https://github.com/XuefengBUPT/IndoorObstacleDiscovery-RG}.

\begin{acknowledgements}
This work was supported by the National Natural Science Foundation of China 62176098, 61703049, the Natural Science Foundation of Hubei Province of China under Grant 2019CFA022,
the national key R\&D program intergovernmental international science and technology innovation cooperation project under Grant No. 2021YFE0101600, the Beijing University of Posts and Telecommunications (BUPT) Excellent Ph.D. Students Foundation under Grant CX2020114.
\end{acknowledgements}

\section*{Appendix}

\begin{appendices}

\section{Principle of Ground-pixel Parallax}
\label{secA1}

In this section,
we state the correctness of Equation \ref{eq:pp2} in the main manuscript,
and prove that this equation determines the relationship between an observed point and the ground.

\begin{figure}[t]
	\centering
	\includegraphics[width=1\linewidth]{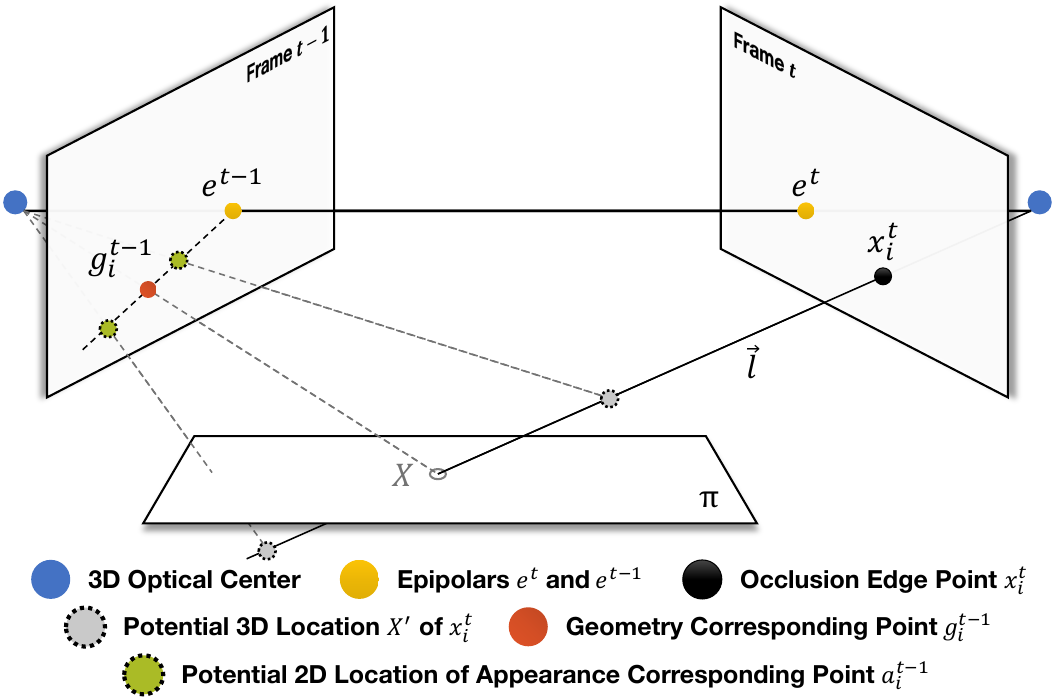}
		\caption{The ground-pixel parallax in two-view geometry.}
	\label{fig:pdp}
\end{figure}

\begin{figure*}
\centering
\includegraphics[width=1\linewidth]{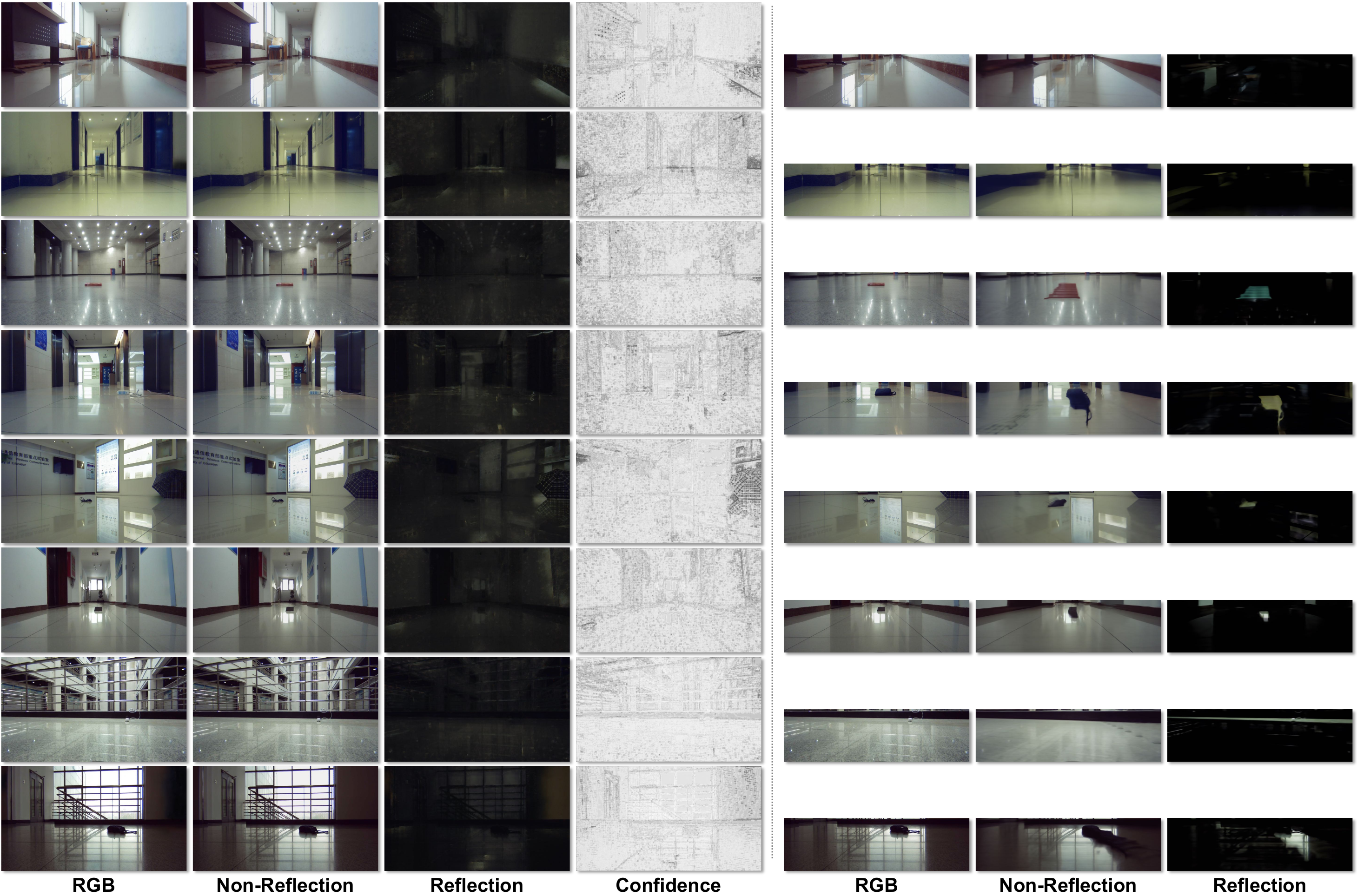}
\caption{Example results of the single frame reflection removal algorithm \cite{lrr} and the multi-frame reflection removal algorithm \cite{nir} on multiple training scenes.}
\label{fig:result_reflectionremoval}
\end{figure*}

First of all,
we give the proof of Equation \ref{eq:pp2} in the main manuscript.
Supposing $I^t$ and $I^{t-1}$ denote two consecutive images from robot's view,
$\pi$ denotes the ground plane,
and $e^t, e^{t-1}$ denote the epipoles on the two images,
which respectively are two intersection points of the two images and a line,
i.e., the line connecting two camera optical centers,
as the yellow points in Fig.\ref{fig:pdp}.
$x^t_i=\{u^t_i,v^t_i,1\}$ denotes an occlusion edge points of image $I^t$ in its homogeneous form,
and it is also the 2D projection of a 3D point $X'$ on image $I^t$.
A ray emitted from $I^t$'s optical center,
denoted as $\vec{l}$,
penetrates $X'$ and $x^t_i$,
and intersects the ground $\pi$ at a 3D point $X$.
In addition,
with the representation of the main manuscript,
the geometric and appearance corresponding points are denoted as $g^{t-1}_i$ and $a^{t-1}_i$.
According to Epipolar Constraints,
the 3D line $\vec{l}$ can be projected to image $I^{t-1}$ to form a projected 2D line,
denoted as $\overrightarrow{g^{t-1}_i e^{t-1}}$.
Since the 3D point $X'$ is on the 3D line $\vec{l}$,
its projection on image $I^{t-1}$,
namely $a^{t-1}_i$,
is on the projected 2D line $\overrightarrow{g^{t-1}_i e^{t-1}}$.
Hence,
these 2D points $a^{t-1}_i$, $g^{t-1}_i$, and $e^{t-1}$ are collinear.
Based on this,
2D point $a^{t-1}_i$ can be stated as:
\begin{equation}
    a^{t-1}_i = g^{t-1}_i + \rho(g^{t-1}_i-e^{t-1})
\end{equation}
where $\rho$ is a scalar.
This equation can be reformulated as Equation \ref{eq:pp2} in the main manuscript.

Then, we discuss the reason that Equation \ref{eq:pp2} in the main manuscript can be used to distinguish the points above the ground and below the ground.
Noteworthily,
since $g^{t-1}_i$ uniquely represents the projection of 3D ground point $X$,
$g^{t-1}_i$ can be considered as the dividing point,
and all points on both sides of this 2D line are partitioned into two spaces,
above the ground and below the ground,
as shown in Fig.\ref{fig:pdp}.
Hence,
the space of point $x^{t}_i$ can be determined by comparing $a^{t-1}_i$ and $g^{t-1}_i$.

\begin{table*}[!tp]
\begin{center}
\caption{Formulation of each feature representing a bounding box.}
\begin{tabular}{@{}lllc@{}}
\toprule
Category    & Feature of bounding box & Formulation  & index      \\
\specialrule{0em}{1pt}{1pt}
\hline
\specialrule{0em}{1pt}{1pt}
\multirow{4}{*}{{\it Edge cue}}    & Max edge response  & $\max{\left(\{x_i^t|x_i^t\in \mathsf{b}\}\right)}$  &  1 \\
& Proportion of most response           & $1/N_j^t\sum_{x_i^t\in \mathsf{b}}\left(x_i^t=\arg\max_r{\sum_{x_i^t\in \mathsf{b}}\left[x_i^t=r\right]}\right) $  & 2  \\
& Average edge response                 & $1/N_j^t\ \sum_{x_i^t\in \mathsf{b}} x_i^t $  & 3  \\
& Average edge response in inner ring   & $1/{\hat{N}}_j^t\sum_{x_i^t\in\check{\mathsf{b}}} x_i^t \,\, \text{, where } \check{\mathsf{b}}=(\mathsf{u}+\mathsf{w}/4,\mathsf{v}+\mathsf{h}/4,\mathsf{w}/2,\mathsf{h}/2)$ &  4 \\
\specialrule{0em}{1pt}{1pt}
\hline
\specialrule{0em}{1pt}{1pt}
\multirow{6}{*}{{\it Pseudo distance}}    &  Normalized area  &  $\left(\mathsf{w}\times \mathsf{h}\right)/\left(\mathsf{W}\times\mathsf{H}\right)$  &  5 \\
    &  Aspect ratio of box  &  $\mathsf{w}/\mathsf{h}$ & 6  \\
    &  X coordinate of the box center  & $\mathsf{u}+\mathsf{w}/2$ & 7 \\
    &  Y coordinate of the box center  & $\mathsf{v}+\mathsf{h}/2$ & 8 \\
    &  Width of box   & $\mathsf{w}$ & 9  \\
    &  Height of box  & $\mathsf{h}$ & 10  \\
    \specialrule{0em}{1pt}{1pt}
\hline
\specialrule{0em}{1pt}{1pt}
\textit{Objectness}  & Occlusion-based objectness score &  Referring to \cite{OLP}  & 11  \\
\specialrule{0em}{1pt}{1pt}
\hline
\specialrule{0em}{1pt}{1pt}
\multirow{6}{*}{{\it Color}}  & Color standard deviation in the H channel  & $\sqrt{1/\mathsf{h}\mathsf{w}\ \sum_{p\in \mathsf{b}}\left(\mathcal{H}\left(p\right)-1/\mathsf{h}\mathsf{w}\sum_{p\in \mathsf{b}}\mathcal{H}\left(p\right)\right)^2} $  & 12  \\
    & Color standard deviation in the S channel  & $\sqrt{1/\mathsf{h}\mathsf{w}\ \sum_{p\in \mathsf{b}}\left(\mathcal{S}\left(p\right)-1/\mathsf{h}\mathsf{w}\sum_{p\in \mathsf{b}}\mathcal{S}\left(p\right)\right)^2}$  & 13  \\
    & Color standard deviation in the V channel  &  $\sqrt{1/\mathsf{h}\mathsf{w}\ \sum_{p\in \mathsf{b}}\left(\mathcal{V}\left(p\right)-1/\mathsf{h}\mathsf{w}\sum_{p\in \mathsf{b}}\mathcal{V}\left(p\right)\right)^2} $ & 14  \\
    & Color contrast in the H channel  & $1-\frac{hist_{\mathsf{b}}^{\mathcal{H}}\cdot hist_{\hat{\mathsf{b}}}^{\mathcal{H}}}{\|hist_{\mathsf{b}}^{\mathcal{H}}\|_{2}\|hist_{\hat{\mathsf{b}}}^{\mathcal{H}}\|_{2}} \text{, where } \hat{\mathsf{b}}=(\mathsf{u}-\mathsf{w}/4,\mathsf{v}-\mathsf{h}/4,2\mathsf{w},2\mathsf{h})$  &  15 \\
    & Color contrast in the S channel  & $1-\frac{hist_{\mathsf{b}}^{\mathcal{S}}\cdot hist_{\hat{\mathsf{b}}}^{\mathcal{S}}}{\|hist_{\mathsf{b}}^{\mathcal{S}}\|_{2}\|hist_{\hat{\mathsf{b}}}^{\mathcal{S}}\|_{2}} \,\,\text{, where } \hat{\mathsf{b}}=(\mathsf{u}-\mathsf{w}/4,\mathsf{v}-\mathsf{h}/4,2\mathsf{w},2\mathsf{h})$  &  16 \\
    & Color contrast in the V channel  & $1-\frac{hist_{\mathsf{b}}^{\mathcal{V}}\cdot hist_{\hat{\mathsf{b}}}^{\mathcal{V}}}{\|hist_{\mathsf{b}}^{\mathcal{V}}\|_{2}\|hist_{\hat{\mathsf{b}}}^{\mathcal{V}}\|_{2}} \,\,\text{, where } \hat{\mathsf{b}}=(\mathsf{u}-\mathsf{w}/4,\mathsf{v}-\mathsf{h}/4,2\mathsf{w},2\mathsf{h})$  &  17 \\
    \specialrule{0em}{1pt}{1pt}
\hline
\specialrule{0em}{1pt}{1pt}
\multirow{2}{*}{{\it Parallax}}  & Homography error  & Referring to Sec. \ref{sec:feat}  & 18  \\
    & Deviation angle  &  Referring to Sec. \ref{sec:feat}  &  19 \\
\bottomrule
\end{tabular}
% }
\label{table:feature_vector}
\end{center}
\vspace{0pt}
\end{table*}

\section{Feasibility of Reflection Removal Methods}
\label{secB}
In scenes of reflection ground, it is intuitive to incorporate reflection removal approaches into the feature extraction part.
Therefore, we employ the state-of-the-art single frame reflection removal algorithm \cite{lrr} (ICCV 2021) and multi-frame reflection removal algorithm \cite{nir} (ECCV 2022), both of which have publicly available code.
Their results on our dataset are visualized in Fig. \ref{fig:result_reflectionremoval}.
Note that since the real-world interfaces the reconstruction of multi-frame-based methods,
we only conduct the reflection removal in the bottom half of the image, i.e., in the ground area.

Observably,
both reflection removal methods are ineffective in removing reflections in our benchmark scenarios.
In fact, they even damage the information of obstacles that needed to be detected. 
The single-frame-based method \cite{lrr} produced confidence maps that failed to highlight the reflection, and the non-reflection images were almost identical to the original RGB images.
Despite being free from the interface of the real world, the multi-frame-based method \cite{nir} still fails to eliminate reflections.
The reason is that \textit{both algorithms require strong textures of main object for adequate reconstruction, but the ground texture is too weak to be perceived}.
In contrast, the reflection has a stronger texture than ground, making it appear as the main object.
Overall, existing reflection removal algorithms cannot be directly used in the scene with reflective ground, and even damage obstacle information.

\begin{figure}
\centering
\includegraphics[width=1\linewidth]{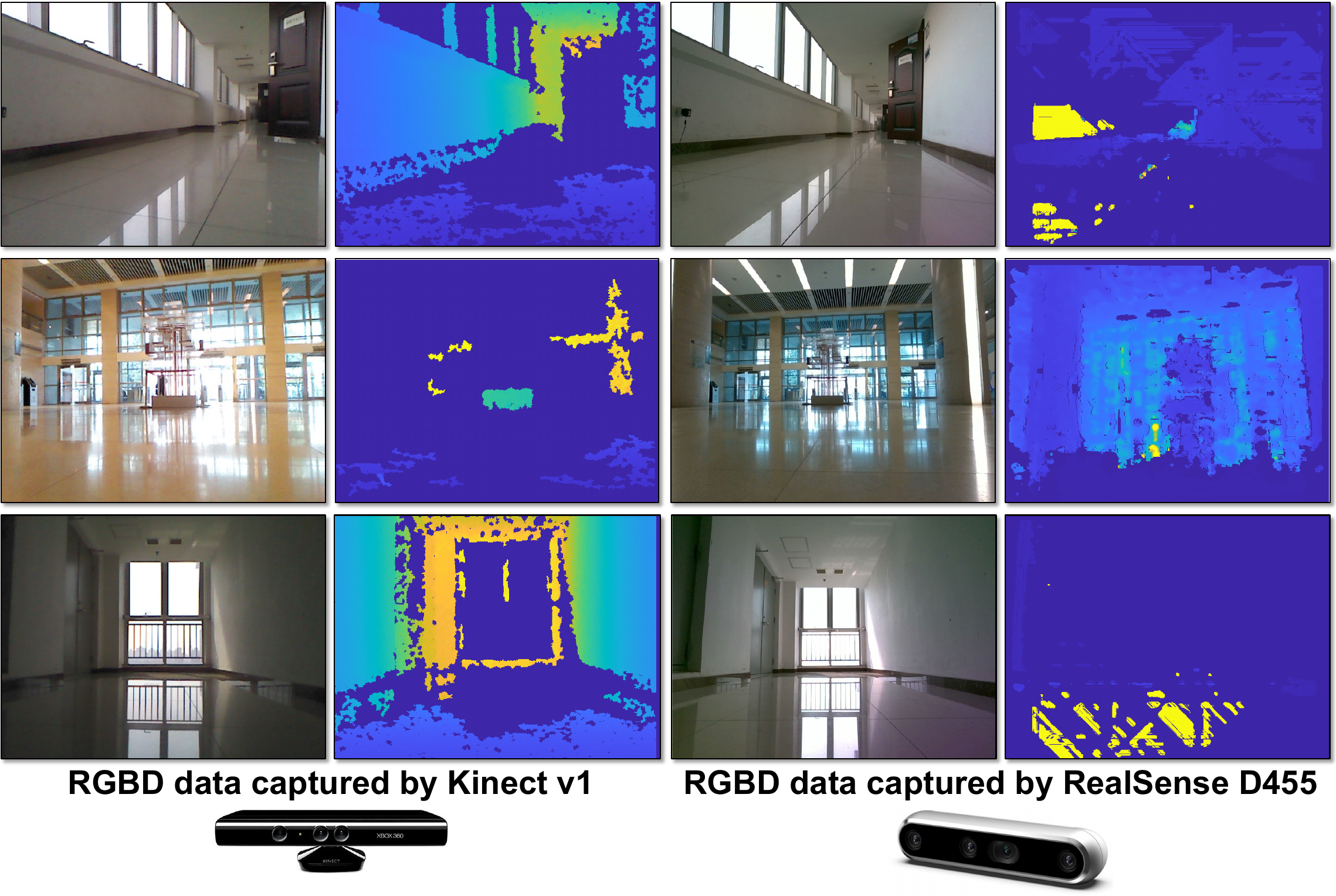}
\caption{RGBD data captured by Kinect v1 and Realsense D455.}
\label{fig:depthsensor}
\end{figure}

\section{Feasibility of Depth Sensors}
\label{secC}
In recent years, multi-modal sensors have been increasingly popular in autonomous driving.
Thus, we evaluate the usability of radar or depth cameras in reflective ground environments.
To this end, we collect depth data of several reflective scenes by two classical sensors,
i.e., the structured light camera (Kinect v1, released in 2010, priced at \$150),
the stereo camera  (RealSense D455, released in 2019, priced at \$249).
The exemplar RGBD data is visualized in Fig. \ref{fig:depthsensor}.
Intuitively, the depth data obtained by these cameras in reflective environments is of such low quality that it cannot be applied in reflective scenes.
Specifically, the structured light camera generates many void areas on the ground plane, while the stereo camera matches corresponding pixels erroneously between the cameras and results in completely incorrect depth data.
Clearly, both types of cameras are unsuitable for reflective ground scenes.

\begin{figure}
\centering
\includegraphics[width=1\linewidth]{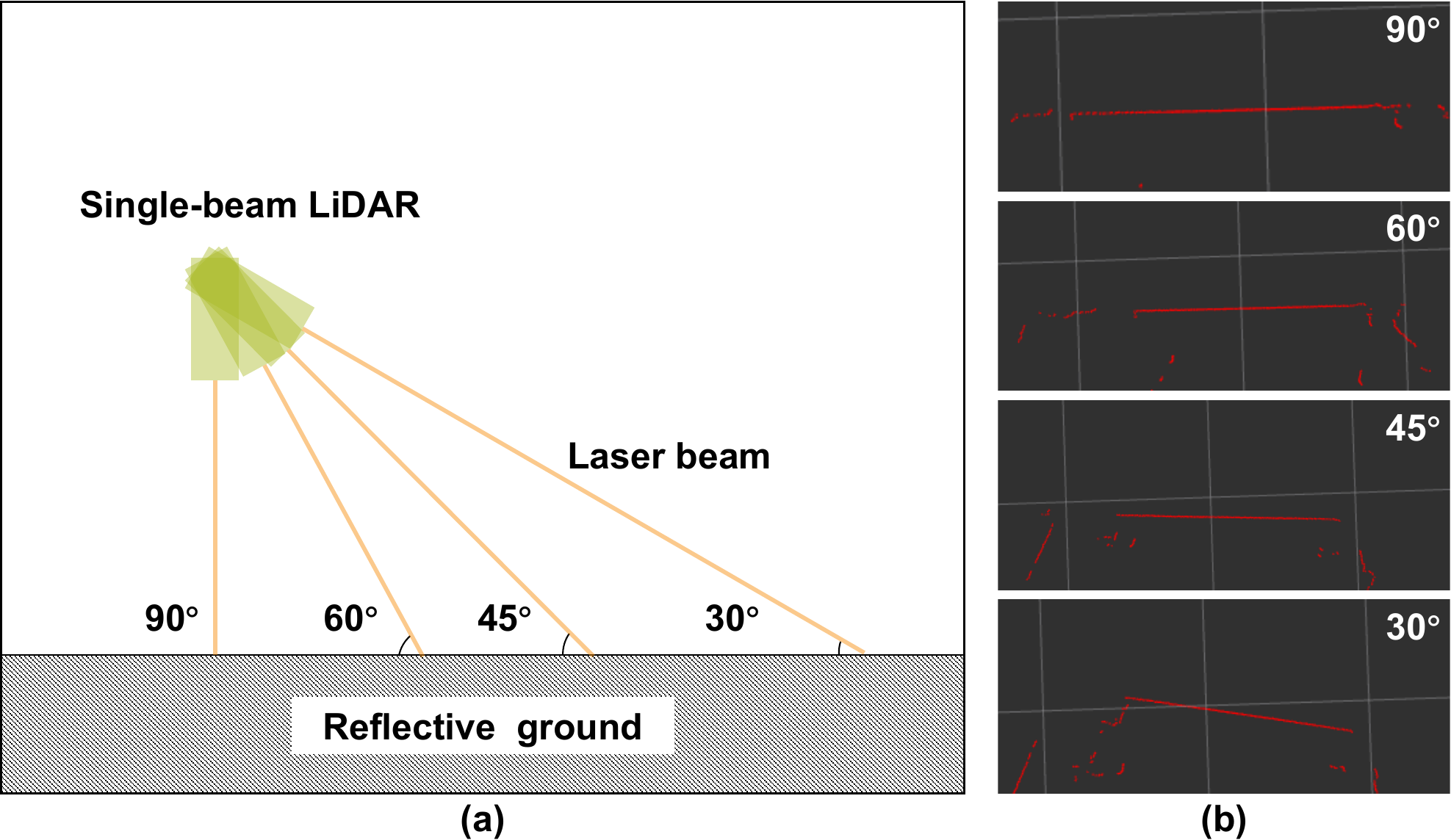}
\caption{Capability tests of laser sensor.
(a) Side view of scanning ground with a single-beam LiDAR.
(b) Laser scan obtained in different angle.}
\label{fig:lidar}
\end{figure}

Furthermore, we conduct capability tests of laser sensor using a single-beam 360-degree LiDAR,
which is illustrated in Fig. \ref{fig:lidar}.
The results illustrate that the single-beam LiDAR is not affected by reflections in all angles,
which means that 3D LiDAR can obtain reliable depth information in reflective scenes.
Unfortunately, although 3D LiDAR almost avoid the issue brought by reflective ground,
it is too expensive to be deployed on a robot compared other depth sensor.

\section{Detailed Formulation of Feature Vector}
\label{secD}
To clearly represent the feature used in this paper,
Table \ref{table:feature_vector} shows the formulation of each feature channel.
Note that, according to Sec. \ref{sec:feat},
$b_j^t$ denotes the $j$-th bounding boxes in the $t$-th image,
and its feature vector is denoted as $v_j^t$,
which consists 19 channels grouped into five categories.
To simplify the notation, we use $\mathsf{b}$ to represent the bounding box $b_j^t$,
specified by its top-left pixel coordinates $\left(\mathsf{u},\mathsf{v}\right)$ and its width and height $\left(\mathsf{w},\mathsf{h}\right)$.
In Table \ref{table:feature_vector}, 
the notation $\check{\mathsf{b}}$ refers to the inner ring of the bounding box $\mathsf{b}$,
while $\hat{\mathsf{b}}$ represents the outer ring.

In the $2$-nd channel,
the notation $\left[.\right]$ represents an indicator function that outputs 1 if the input is correct and 0 otherwise.
In the $5$-th channel,
$\left(\mathsf{W},\mathsf{H}\right)$ denotes the width and height of the input image.
For the $12$-th - $17$-th channels,
the variables $\mathcal{H}$, $\mathcal{S}$, and $\mathcal{V}$ correspond to the input image’s HSV channels,
and $\mathcal{H}\left(p\right)$ denotes the $p$-th pixel in the channel $\mathcal{H}$.
Note that, the color contrast formulation involves the normalized histogram of box $\mathsf{b}$’s H channel,
represented as $hist_\mathsf{b}^\mathcal{H}$,
which is discretized into 18 bins denoted by $\{h_k^\mathcal{H},k\in\left[1,18\right]\}$,
where $k$ ranges from 1 to 18.
The value of $h_k^\mathcal{H}$ is obtained as the sum over all pixels $p$ in box $\mathsf{b}$ such that $h_k^\mathcal{H}=\sum_{p\in \mathsf{b}}\left[\left\lfloor\frac{\mathcal{H}\left(p\right)}{360/18}\right\rfloor=k\right]$.
Additionally, similar histograms $hist_\mathsf{b}^\mathcal{S}$, $hist_\mathsf{b}^\mathcal{V}$, $hist_{\hat{\mathsf{b}}}^\mathcal{H}$, $hist_{\hat{\mathsf{b}}}^\mathcal{S}$, $hist_{\hat{\mathsf{b}}}^\mathcal{V}$ are computed in the same way.

\end{appendices}

\balance
\bibliographystyle{spmpsci}
\bibliography{ORG_Final}

\begin{thebibliography}{10}
\providecommand{\url}[1]{{#1}}
\providecommand{\urlprefix}{URL }
\expandafter\ifx\csname urlstyle\endcsname\relax
  \providecommand{\doi}[1]{DOI~\discretionary{}{}{}#1}\else
  \providecommand{\doi}{DOI~\discretionary{}{}{}\begingroup
  \urlstyle{rm}\Url}\fi

\bibitem{gPb}
{Arbelaez}, P., {Maire}, M., {Fowlkes}, C., {Malik}, J.: Contour detection and
  hierarchical image segmentation.
\newblock IEEE Transactions on Pattern Analysis and Machine Intelligence
  (TPAMI) \textbf{33}(5), 898--916 (2011)

\bibitem{Bian2021wb}
Bian, J.W., Zhan, H., Wang, N., Li, Z., Zhang, L., Shen, C., Cheng, M.M., Reid,
  I.: Unsupervised scale-consistent depth learning from video.
\newblock International Journal of Computer Vision (IJCV) \textbf{129}(9),
  2548--2564 (2021)

\bibitem{2011Stereo}
Broggi, A., Buzzoni, M., Felisa, M., Zani, P.: Stereo obstacle detection in
  challenging environments: The {VIAC} experience.
\newblock In: IEEE/RSJ International Conference on Intelligent Robots and
  Systems (IROS) (2011)

\bibitem{DeepLab}
Chen, L.C., Zhu, Y., Papandreou, G., Schroff, F., Adam, H.: Encoder-decoder
  with atrous separable convolution for semantic image segmentation.
\newblock In: European Conference on Computer Vision (ECCV) (2018)

\bibitem{Chen2009vc}
Chen, T., Vemuri, B.C., Rangarajan, A., Eisenschenk, S.J.: Group-wise point-set
  registration using a novel cdf-based havrda-charv{\'a}t divergence.
\newblock International Journal of Computer Vision (IJCV) \textbf{86}(1), 111
  (2009)

\bibitem{EM}
{Conrad}, D., {DeSouza}, G.N.: Homography-based ground plane detection for
  mobile robot navigation using a modified em algorithm.
\newblock In: IEEE International Conference on Robotics and Automation (ICRA)
  (2010)

\bibitem{DF}
Criminisi, A., Shotton, J., Konukoglu, E.: Decision Forests: A Unified
  Framework for Classification, Regression, Density Estimation, Manifold
  Learning and Semi-Supervised Learning.
\newblock Now Publishers Inc (2012)

\bibitem{lrr}
Dong, Z., Xu, K., Yang, Y., Bao, H., Xu, W., Lau, R.W.: Location-aware single
  image reflection removal.
\newblock In: IEEE/CVF International Conference on Computer Vision (ICCV)
  (2021)

\bibitem{CalibUrl}
Dongfu, Z., Zheng, C.: {The code of CamOdomCalibraTool}.
\newblock \url{https://github.com/MegviiRobot/CamOdomCalibraTool}

\bibitem{Ghodrati2017ve}
Ghodrati, A., Diba, A., Pedersoli, M., Tuytelaars, T., Van~Gool, L.:
  Deepproposals: Hunting objects and actions by cascading deep convolutional
  layers.
\newblock International Journal of Computer Vision (IJCV) \textbf{124}(2),
  115--131 (2017)

\bibitem{MergeNet}
{Gupta}, K., {Javed}, S.A., {Gandhi}, V., {Krishna}, K.M.: Mergenet: A deep net
  architecture for small obstacle discovery.
\newblock In: IEEE International Conference on Robotics and Automation (ICRA)
  (2018)

\bibitem{MV}
Hartley, R., Zisserman, A.: Multiple View Geometry in Computer Vision.
\newblock Cambridge University Press (2003)

\bibitem{He_2016_CVPR}
He, K., Zhang, X., Ren, S., Sun, J.: Deep residual learning for image
  recognition.
\newblock In: IEEE Conference on Computer Vision and Pattern Recognition (CVPR)
  (2016)

\bibitem{Heng2013CamOdoCal}
Heng, L., Bo, L., Pollefeys, M.: Camodocal: Automatic intrinsic and extrinsic
  calibration of a rig with multiple generic cameras and odometry.
\newblock In: IEEE/RSJ International Conference on Intelligent Robots and
  Systems (IROS) (2013)

\bibitem{Hoiem2011te}
Hoiem, D., Efros, A.A., Hebert, M.: Recovering occlusion boundaries from an
  image.
\newblock International Journal of Computer Vision (IJCV) \textbf{91}(3),
  328--346 (2011)

\bibitem{Hua_ICCVW}
{Hua}, M., {Nan}, Y., {Lian}, S.: Small obstacle avoidance based on rgb-d
  semantic segmentation.
\newblock In: IEEE International Conference on Computer Vision Workshop (ICCVW)
  (2019)

\bibitem{Jia_2007}
{Jia}, J.: Single image motion deblurring using transparency.
\newblock In: IEEE Conference on Computer Vision and Pattern Recognition (CVPR)
  (2007)

\bibitem{Kalal2010Forward}
Kalal, Z., Mikolajczyk, K., Matas, J.: Forward-backward error: Automatic
  detection of tracking failures.
\newblock In: IEEE International Conference on Pattern Recognition (ICPR)
  (2010)

\bibitem{adam}
Kingma, D., Ba, J.: Adam: A method for stochastic optimization.
\newblock In: International Conference on Learning Representations (ICLR)
  (2015)

\bibitem{alexnet}
Krizhevsky, A., Sutskever, I., Hinton, G.E.: Imagenet classification with deep
  convolutional neural networks.
\newblock In: Advances in Neural Information Processing Systems (NIPS) (2012)

\bibitem{Markov}
{Kumar}, S., {Karthik}, M.S., {Krishna}, K.M.: Markov random field based small
  obstacle discovery over images.
\newblock In: IEEE International Conference on Robotics and Automation (ICRA)
  (2014)

\bibitem{Li2019wt}
Li, H., Liu, Y., Ouyang, W., Wang, X.: Zoom out-and-in network with map
  attention decision for region proposal and object detection.
\newblock International Journal of Computer Vision (IJCV) \textbf{127}(3),
  225--238 (2019)

\bibitem{Lin}
{Lin}, C., {Jiang}, S., {Yueh-Ju Pu}, {Song}, K.: Robust ground plane detection
  for obstacle avoidance of mobile robots using a monocular camera.
\newblock In: IEEE/RSJ International Conference on Intelligent Robots and
  Systems (IROS) (2010)

\bibitem{Lindeberg1998tl}
Lindeberg, T.: Edge detection and ridge detection with automatic scale
  selection.
\newblock International Journal of Computer Vision (IJCV) \textbf{30}(2),
  117--156 (1998)

\bibitem{Lis}
Lis, K., Nakka, K.K., Fua, P., Salzmann, M.: Detecting the unexpected via image
  resynthesis.
\newblock In: IEEE/CVF International Conference on Computer Vision (ICCV)
  (2019)

\bibitem{OFNet}
Lu, R., Xue, F., Zhou, M., Ming, A., Zhou, Y.: Occlusion-shared and
  feature-separated network for occlusion relationship reasoning.
\newblock In: IEEE/CVF International Conference on Computer Vision (ICCV)
  (2019)

\bibitem{LK}
Lucas, B.D., Kanade, T.: An iterative image registration technique with an
  application to stereo vision.
\newblock In: International Joint Conference on Artificial Intelligence (IJCAI)
  (1981)

\bibitem{OLP}
Ma, J., Ming, A., Huang, Z., Wang, X., Zhou, Y.: Object-level proposals.
\newblock In: IEEE International Conference on Computer Vision (ICCV) (2017)

\bibitem{DecompH}
Malis, E., Vargas, M.: Deeper understanding of the homography decomposition for
  vision-based control.
\newblock Research Report RR-6303, INRIA (2007)

\bibitem{JMOD2}
Mancini, M., Costante, G., Valigi, P., Ciarfuglia, T.A.: {J-MOD2}: Joint
  monocular obstacle detection and depth estimation.
\newblock IEEE Robotics and Automation Letters (RA-L) \textbf{3}(3), 1--1
  (2018)

\bibitem{IS}
Ming, A., Wu, T., Ma, J., Sun, F., Zhou, Y.: Monocular depth-ordering reasoning
  with occlusion edge detection and couple layers inference.
\newblock In: IEEE Intelligent Systems (IS), vol.~31, pp. 54--65 (2016)

\bibitem{ICIP15}
{Ming}, A., {Xun}, B., {Ni}, J., {Gao}, M., {Zhou}, Y.: Learning discriminative
  occlusion feature for depth ordering inference on monocular image.
\newblock In: IEEE International Conference on Image Processing (ICIP) (2015)

\bibitem{nir}
Nam, S., Brubaker, M.A., Brown, M.S.: Neural image representations for
  multi-image fusion and layer separation.
\newblock In: S.~Avidan, G.~Brostow, M.~Ciss{\'e}, G.M. Farinella, T.~Hassner
  (eds.) Europe Conference on Computer Vision (ECCV) (2022)

\bibitem{Panahandeh}
{Panahandeh}, G., {Jansson}, M.: Vision-aided inertial navigation based on
  ground plane feature detection.
\newblock IEEE/ASME Transactions on Mechatronics (TMECH) \textbf{19}(4),
  1206--1215 (2014)

\bibitem{ENet}
Paszke, A., Chaurasia, A., Kim, S., Culurciello, E.: Enet: A deep neural
  network architecture for real-time semantic segmentation.
\newblock In: International Conference on Learning Representations (ICLR)
  (2017)

\bibitem{LAF}
Pinggera, P., Ramos, S., Gehrig, S., Franke, U., Rother, C., Mester, R.: Lost
  and found: detecting small road hazards for self-driving vehicles.
\newblock In: IEEE/RSJ International Conference on Intelligent Robots and
  Systems (IROS) (2016)

\bibitem{UON}
{Ramos}, S., {Gehrig}, S., {Pinggera}, P., {Franke}, U., {Rother}, C.:
  Detecting unexpected obstacles for self-driving cars: Fusing deep learning
  and geometric modeling.
\newblock In: IEEE Intelligent Vehicles Symposium (IV) (2017)

\bibitem{Saxena2008tc}
Saxena, A., Chung, S.H., Ng, A.Y.: 3-d depth reconstruction from a single still
  image.
\newblock International Journal of Computer Vision (IJCV) \textbf{76}(1),
  53--69 (2008)

\bibitem{ICP}
Sharp, G., Lee, S., Wehe, D.: {ICP} registration using invariant features.
\newblock IEEE Transactions on Pattern Analysis and Machine Intelligence
  (TPAMI) \textbf{24}(1), 90--102 (2002)

\bibitem{FCN}
{Shelhamer}, E., {Long}, J., {Darrell}, T.: Fully convolutional networks for
  semantic segmentation.
\newblock IEEE Transactions on Pattern Analysis and Machine Intelligence
  (TPAMI) \textbf{39}(4), 640--651 (2017)

\bibitem{2020LiDAR}
Singh, A., Kamireddypalli, A., Gandhi, V., Krishna, K.M.: Lidar guided small
  obstacle segmentation.
\newblock In: IEEE/RSJ International Conference on Intelligent Robots and
  Systems (IROS) (2020)

\bibitem{9134735}
{Sun}, L., {Yang}, K., {Hu}, X., {Hu}, W., {Wang}, K.: Real-time fusion network
  for rgb-d semantic segmentation incorporating unexpected obstacle detection
  for road-driving images.
\newblock IEEE Robotics and Automation Letters (RA-L)  (2020)

\bibitem{Xie2017tc}
Xie, S., Tu, Z.: Holistically-nested edge detection.
\newblock International Journal of Computer Vision (IJCV) \textbf{125}(1),
  3--18 (2017)

\bibitem{depthpr}
Xue, F., Cao, J., Zhou, Y., Sheng, F., Wang, Y., Ming, A.: Boundary-induced and
  scene-aggregated network for monocular depth prediction.
\newblock Pattern Recognition (PR) \textbf{115}, 107901 (2021)

\bibitem{ICRA}
{Xue}, F., {Ming}, A., {Zhou}, M., {Zhou}, Y.: A novel multi-layer framework
  for tiny obstacle discovery.
\newblock In: IEEE International Conference on Robotics and Automation (ICRA)
  (2019)

\bibitem{tip}
{Xue}, F., {Ming}, A., {Zhou}, Y.: Tiny obstacle discovery by occlusion-aware
  multilayer regression.
\newblock IEEE Transactions on Image Processing (TIP) \textbf{29}, 9373--9386
  (2020)

\bibitem{BiSeNet}
Yu, C., Wang, J., Peng, C., Gao, C., Yu, G., Sang, N.: Bisenet: Bilateral
  segmentation network for real-time semantic segmentation.
\newblock In: European Conference on Computer Vision (ECCV) (2018)

\bibitem{JinICRA}
{Zhou}, J., {Li}, B.: Homography-based ground detection for a mobile robot
  platform using a single camera.
\newblock In: IEEE International Conference on Robotics and Automation (ICRA)
  (2006)

\bibitem{JinICIP}
{Zhou}, J., {Li}, B.: Robust ground plane detection with normalized homography
  in monocular sequences from a robot platform.
\newblock In: IEEE International Conference on Image Processing (ICIP) (2006)

\bibitem{icipTracker}
Zhou, M., Ma, J., Ming, A., Zhou, Y.: Objectness-aware tracking via
  double-layer model.
\newblock In: IEEE International Conference on Image Processing (ICIP) (2018)

\bibitem{ijcv}
Zhou, Y., Bai, X., Liu, W., Latecki., L.: Similarity fusion for visual
  tracking.
\newblock International Journal of Computer Vision (IJCV) \textbf{118}(3),
  337--363 (2016)

\end{thebibliography}

\end{sloppypar}
\end{document}